\documentclass[11pt]{article}

\usepackage[final]{acl}

\newif\ifcomment\commenttrue

\usepackage[a-1b]{pdfx}

\usepackage{framed}
\usepackage{mdwlist}
\usepackage{siunitx}
\usepackage{latexsym}
\usepackage{colortbl}
\usepackage{xcolor}
\usepackage{nicefrac}
\usepackage{booktabs}
\usepackage{fnpct}
\usepackage{amsfonts}
\usepackage[T1]{fontenc}
\usepackage{bold-extra}
\usepackage{amsmath}
\usepackage{amssymb}
\usepackage{bm}
\usepackage{graphicx}
\usepackage{mathtools}
\usepackage{microtype}
\usepackage{multirow}
\usepackage{multicol}
\usepackage{xpatch}
\usepackage{latexsym,comment}
\usepackage[normalem]{ulem}

\newcommand*{\missingreference}{{\Huge \colorbox{red}{?reference?}}}
\newcommand*{\missingcitation}{{\Huge \colorbox{red}{?citation?}}}

\makeatletter
\xpatchcmd{\@setref}{\bfseries}{\missingreference}{}{}
\def\@citex[#1]#2{\leavevmode
    \let\@citea\@empty
    \@cite{\@for\@citeb:=#2\do
        {\@citea\def\@citea{,\penalty\@m\ }%
            \edef\@citeb{\expandafter\@firstofone\@citeb\@empty}%
            \if@filesw\immediate\write\@auxout{\string\citation{\@citeb}}\fi
            \@ifundefined{b@\@citeb}{\hbox{\reset@font\missingcitation}%
                \G@refundefinedtrue
                \@latex@warning
                {Citation `\@citeb' on page \thepage \space undefined}}%
            {\@cite@ofmt{\csname b@\@citeb\endcsname}}}}{#1}}
\makeatother

\newcommand{\gem}[1]{\mbox{\textsc{gem}}}
\newcommand{\abr}[1]{\textsc{#1}}

\newcommand{\g}{\, | \,}

\newcommand{\hidetext}[1]{}
\newcommand{\ignore}[1]{}
\newif\ifcomment
\commenttrue

\ifcomment
    \newcommand{\pinaforecomment}[3]{\colorbox{#1}{\parbox{.8\linewidth}{#2: #3}}}

    \newcommand{\prtodo}[1]{\pinaforecomment{lightblue}{pr}{#1}}
    \newcommand{\prtodoi}[1]{\pinaforecomment{lightblue}{pr}{#1}}
\else
    \newcommand{\pinaforecomment}[3]{}
    \newcommand{\prtodo}[1]{}
    \newcommand{\prtodoi}[1]{}
\fi

\newcommand{\smallurl}[1]{ \begin{tiny}\url{#1}\end{tiny}}

\definecolor{lightblue}{HTML}{3cc7ea}
\definecolor{CUgold}{HTML}{CFB87C}
\definecolor{grey}{rgb}{0.95,0.95,0.95}
\definecolor{ceil}{rgb}{0.57, 0.63, 0.81}
\definecolor{UMDred}{HTML}{ed1c24}
\definecolor{UMDyellow}{HTML}{ffc20e}

\usepackage{times}
\usepackage{latexsym}
\usepackage{booktabs}
\usepackage{subcaption}
\usepackage{multirow}
\usepackage[table]{xcolor}
\usepackage[ruled,vlined,linesnumbered]{algorithm2e}

\usepackage{tcolorbox}
\usepackage{booktabs}
\usepackage{xcolor}
\usepackage{array}
\usepackage{tabularx}
\tcbuselibrary{skins, breakable}

\definecolor{SIcolor}{HTML}{F5C6C6}
\definecolor{CQcolor}{HTML}{FDE2A7}
\definecolor{AQcolor}{HTML}{B8F0CD}
\definecolor{RQcolor}{HTML}{B3D9F2}
\definecolor{NOcolor}{HTML}{D5D8DC}

\usepackage{xcolor}

\definecolor{aqbg}{HTML}{FCF3DC} %
\definecolor{sibg}{HTML}{D8E9F7} %
\definecolor{cqbg}{HTML}{DFF2E5} %
\definecolor{rqbg}{HTML}{FBDEDB} %
\definecolor{nobg}{HTML}{EFE3F5} %

\definecolor{aqfg}{HTML}{7A5C00} %
\definecolor{sifg}{HTML}{1B4F72} %
\definecolor{cqfg}{HTML}{145A32} %
\definecolor{rqfg}{HTML}{922B21} %
\definecolor{nofg}{HTML}{4A235A} %

\makeatletter
\newcommand{\defineact}[2]{\@namedef{act@#1}{#2}} %
\newcommand{\discoact}[1]{%
  \ifcsname act@#1\endcsname
    {\setlength{\fboxsep}{1.5pt}%
     \colorbox{\@nameuse{act@#1}bg}{\abr{#1}}}%
  \else
    \PackageWarning{act}{Unknown act: #1}\abr{#1}%
  \fi}
\makeatother

\defineact{Provide Background}{aq}
\defineact{Provide Example}{aq}
\defineact{Assert Answer}{aq}
\defineact{Provide Reasoning}{aq}
\defineact{Provide Reasoning or Justification}{aq}
\defineact{Provide Recommendation}{aq}
\defineact{Answer Question outside of Interpretation Space}{aq}
\defineact{Cite External Source}{aq}
\defineact{Clarification}{si}
\defineact{Probe Context}{si}
\defineact{Counter-Question}{si}
\defineact{Reject Presupposition}{cq}
\defineact{Correct Fact/Term}{cq}
\defineact{Comment on Question}{cq}
\defineact{Redirect Formulation}{cq}
\defineact{Disambiguate or Surface Interpretations}{cq}
\defineact{Surface Interpretations}{cq}

\defineact{Direct to Resource}{rq}
\defineact{Recommend Expert}{rq}
\defineact{Non-Answer}{no}
\defineact{Presentational}{no}
\defineact{Aside/Other Remark}{no}
\defineact{Offer}{no}

\usepackage[T1]{fontenc}

\usepackage[utf8]{inputenc}

\usepackage{microtype}

\usepackage{inconsolata}

\usepackage{graphicx}

\title{\raisebox{-0.3em}{\includegraphics[height=1.4em]{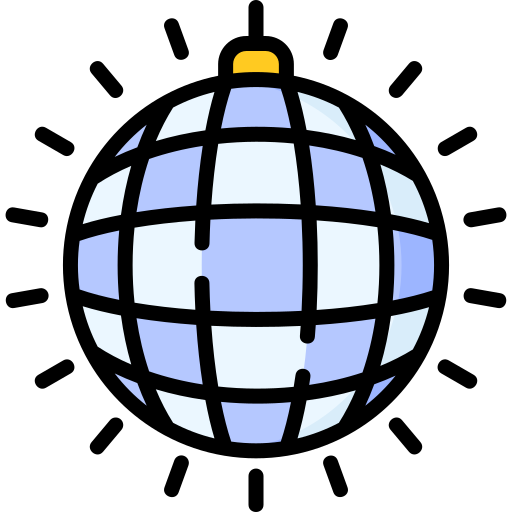}}~\abr{DiscoTrace}: Representing and Comparing Answering Strategies of Humans and LLMs in Information-Seeking Question Answering}

\newcommand{\cs}{$^{\diamondsuit}$}
\newcommand{\ntu}{$^{\spadesuit}$}
\newcommand{\umiacs}{$^{\clubsuit}$}

\author{%
    \textbf{Neha Srikanth}\cs, 
    \textbf{Jordan Boyd-Graber} \ntu, 
    \textbf{Rachel Rudinger}\cs \umiacs\\
    \cs{}Computer Science, University of Maryland \\
    \umiacs{}UMIACS, University of Maryland \\
    \ntu{}Computing and Data Science, NTU Singapore \\
    \texttt{nehasrik@umd.edu}
}

\newcommand{\dt}{\abr{DiscoTrace}}
\newcommand{\strategy}{$S_{DT}$}

\newcommand{\model}[1]{\texttt{\small #1}}
\newcommand{\subreddit}[1]{\texttt{\small #1}}
\newcommand{\annotateinterp}{\includegraphics[height=8pt,width=8pt]{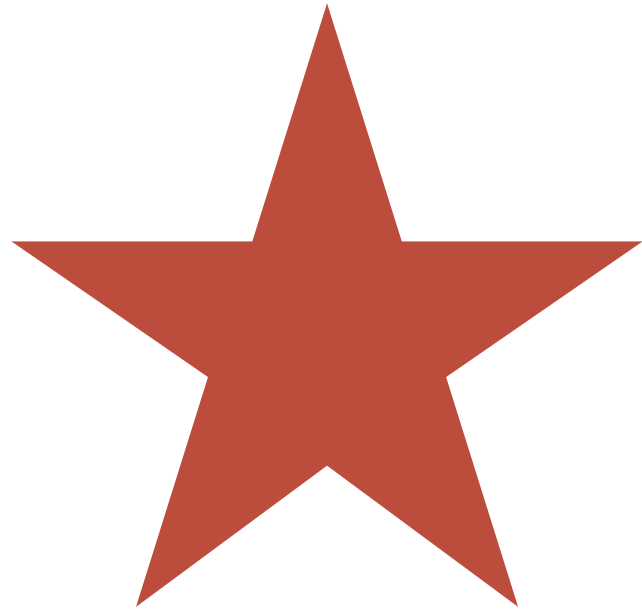}}

\newtcolorbox[list inside=prompt,auto counter,number within=section]{prompt}[1][]{
    breakable,
    colbacktitle=black!60,
    fonttitle=\small,
    coltitle=white,
    fontupper=\footnotesize,
    boxsep=4pt,
    left=0pt,
    right=0pt,
    top=0pt,
    bottom=0pt,
    boxrule=1pt,
    #1,
}

\begin{document}
\maketitle
\begin{abstract}
We introduce \dt{}, a method to identify the rhetorical strategies  answerers use when responding to information-seeking questions.
\dt{} represents answers as a sequence of question-related discourse acts paired with interpretations of the original question, annotated on top of rhetorical structure theory parses.
Applying \dt{} to answers from nine different  communities reveals that communities have diverse preferences for answer construction.
In contrast, LLMs do not exhibit rhetorical diversity in their answers, even when prompted to mimic specific human community answering guidelines.
LLMs also systematically opt for breadth, addressing interpretations of questions that human answerers choose not to address. 
The rich, community-sensitive answering behavior structurally revealed by \dt{} can guide the development of \textit{pragmatic} LLM answerers that are more attuned to contextual information needs.

\end{abstract}

\section{Introduction}

Human--AI interactions have moved beyond simple fact retrieval (``\textit{What is the capital of France?''}) to more complex questions~\cite{xu2022we} requiring longer-form responses~\cite{fan2019eli5}, such as asking for subjective interpretation of facts, empathy-seeking, or even advice~\cite{shelby2025taxonomy, chatterji2025people}, more closely resembling the questions humans may often ask of each other.
These questions may even be
ambiguous~\cite{stelmakh-etal-2022-asqa, cole-etal-2023-selectively},
vague, or contain false assumptions~\cite{kim2021linguist, srikanth-etal-2024-pregnant}.
Humans or AI can adopt different strategies when answering questions: ask a follow-up question, probe for more context, or make their best guess about the information needs of the user~\cite{srikanth2025no}.
As users turn to AI for questions that may be complex,
flawed, domain-specific, or multifaceted, it is imperative that we understand \textit{what} makes one answer better than another.\footnote{Code and data released at \url{https://github.com/nehasrikn/discotrace}.}

\begin{figure}[t!]
\centering
\includegraphics[scale=0.65]{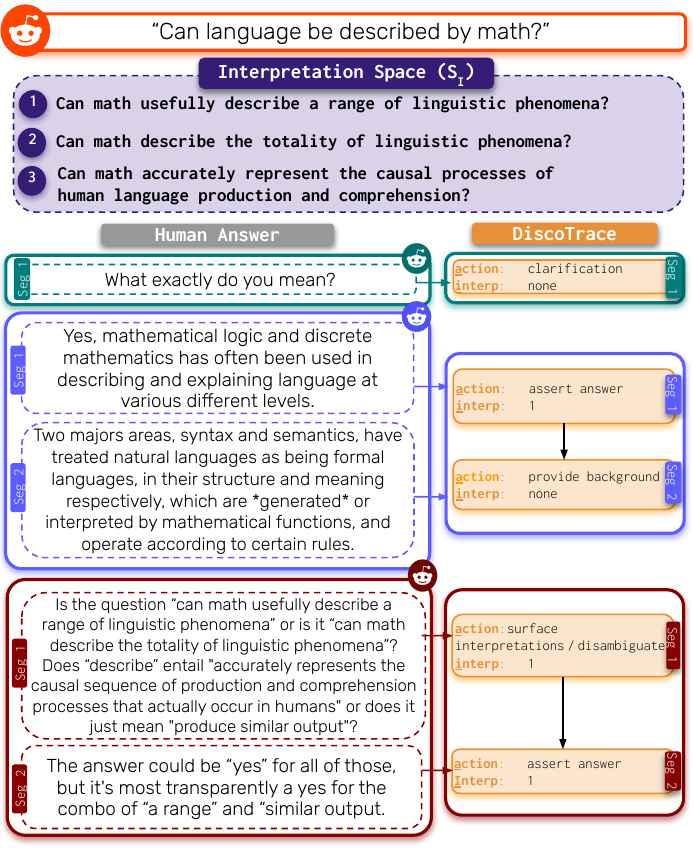}
\caption{Given a question (top) and its interpretations (middle), \dt{} provides a scaffold (bottom right) of answer rhetorical structure (bottom left) with a sequence of tuples of discourse acts and interpretations.}
\label{fig:teaser-example}
\end{figure}

While reinforcement learning from human feedback (RLHF) and human preference judgments~\cite{pmlr-v162-ethayarajh22a} aim to make answers more useful, they \textit{implicitly} conflate many factors---factuality, completeness, style, length, relevance, or tone---with \textit{rhetorical strategy}~\cite{hu-etal-2023-decipherpref}: the choices an answerer makes about \textit{what} to address and \textit{how} to structure a response.
From a practical perspective, this diffuse signal makes it hard
for models and researchers to learn \textit{what} about an answer makes it good or how to generalize to use this answer to help answer a different question.
From a theoretical perspective, we as researchers lack the tools to
model \textit{rhetorical} differences, both in \textit{content}
(\textit{what was addressed?}) and in \textit{strategy} (\textit{how
was it addressed?}), in answers produced by humans
in the same community, humans across communities, or LLMs.

We introduce \dt~(\S\ref{sec:discotrace}), a discourse-based representation of a long-form answer to an information-seeking question.
An answer \dt{} consists of a sequence of question-related \textit{discourse acts}~\cite{searle1969speech}---communicative moves (e.g., question an assumption or provide background)---each paired with specific \textit{interpretations} of the original question (e.g., this is asking about Georgia the American state, not the country), annotated on top of rhetorical structure theory parses~\cite[RST]{mann1987rhetorical}.
The resulting sequence reveals the answerer's \textit{strategy}, providing a high-level, interpretable mechanism to compare answers: which interpretations were covered,
whether additional information was sought, etc.
Figure~\ref{fig:teaser-example} shows a
question taken from a QA community on linguistics (\subreddit{r/asklinguistics}).
The \dt{} of each answer surfaces three different answering strategies: the first answerer requests clarification, the second adopts an interpretation and answers it, and the third surfaces different readings of the question.

\dt{} reveals just how differently humans answer questions in different online communities (\S\ref{sec:communities}) with different preferences and objectives~\cite{kumar-etal-2025-compo}, 
which LLMs fail to do even when they are prompted to mimic answers from a particular community. 
Moreover, LLMs take a shallower approach to QA, hedging over many interpretations, unlike humans, who favor deeply answering only a few (\S\ref{sec:interpretations}).
This extends prior observations that LLM text is lengthier and less
lexically and grammatically diverse than human
text~\cite{poddar2025brevity, reinhart2025llms, munoz2024contrasting}:
\textbf{\dt{} reveals that the homogeneity persists at a higher level of
abstraction, in the strategic discourse choices LLMs make when constructing an answer to questions from vastly different domains.}
Consequently, \dt{} can help us develop \textbf{rhetorically flexible, pragmatic LLMs that adapt their strategies when answering questions} to diverse, sometimes competing, preferences or objectives---e.g., being more empathetic, more skeptical, or more
evidence-based as context demands.

\section{Motivation and Background}
\label{sec:background}
Unlike factoid questions that can be answered within a few tokens and typically have a single correct answer~\cite{kwiatkowski-etal-2019-natural}, \textit{long form} question-answering (LFQA) involves questions that may be open-ended, subjective, or complex~\cite{dietz2017trec, fan2019eli5}.
Evaluating answers to these types of questions is challenging, since
there are often many ways to respond, and \textbf{answerers may employ fundamentally different---but \textit{valid}---strategies
when responding to a question}~\cite{rogers2023qa}.
The choice of strategy in an answer may be driven by the answerer's
expertise~\cite{malaviya-etal-2024-expertqa}, the context in which the
question is asked, the answerer's reasoning or understanding of the
user's information need~\cite{taylor1962process, srikanth2025no},
and the question
type~\cite{cao-wang-2021-controllable}, etc.
\textbf{These factors admit a panoply of diverse answers with distinct rhetorical strategies.}

Much LFQA research has been centered on the ELI5 dataset~\cite{fan2019eli5, krishna2021hurdles} from the subreddit \subreddit{r/explainlikeimfive}, a community that rewards conceptual
simplicity in explanations, but this is just \textit{one example} of
an objective that answerers may choose to prioritize~\cite{adamic2008knowledge}.
Much work on learning from human preferences~\cite{ouyang2022training}
also tries to align models with the preferences from
a \textit{prototypical} user, but this can similarly collapse the
real-world diversity in objectives that may drive heterogeneous
preferences~\cite{bakker2022fine}.

\begin{figure*}
\centering
\includegraphics[scale=0.65]{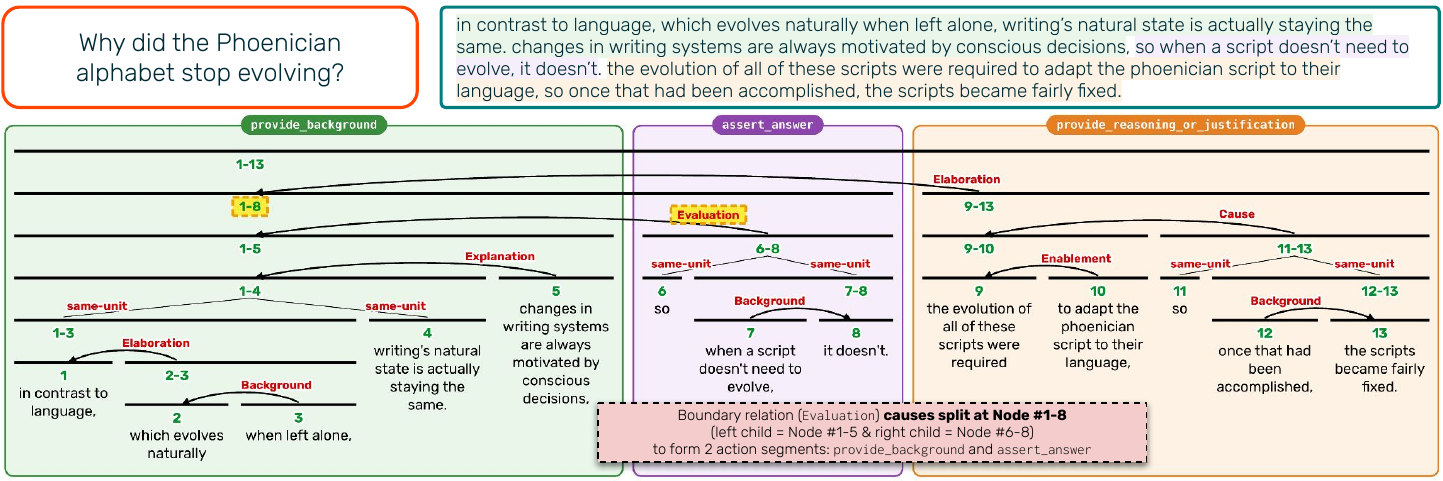}
\caption{Rhetorical structure theory (RST) trees, designed to explain text structure by identifying relationship between text subsegments, provide useful discourse scaffolding, but not enough information alone to characterize high-level answering strategy. We first pre-process answers with RST, identifying \textit{boundary} nodes, where children likely belong to different discourse acts and their subtrees define segment boundaries. Then, using the procedure in \S\ref{subsec:rst}, we construct discourse segments which are then automatically labeled with answer-oriented discourse acts.}
\label{fig:rst-parsing}
\end{figure*}

Given the complexity of the task, \citet{xu2023critical} argue
for \textit{multifaceted} evaluation of answers, encouraging the use
of finer-grained metrics, noting that in some cases, experts
themselves were at odds in their preference judgments.
In this spirit, we study answers and human preferences in diverse communities from a \textit{rhetorical} perspective,
characterizing their discourse \textit{structure}~\cite{xu2022we} and
choice of question interpretation.

\paragraph{Discourse Structure.} Speakers use language to accomplish communicative goals~\cite{searle2014speech}.
A \textit{discourse act} is a functional label that helps characterize
the communicative role that an utterance or a piece of text plays with
respect to its context: the same surface form can perform different acts
depending on the question it responds to. 
An explanation like
\textit{``because the liver already filters toxins continuously''}
justifies an answer to \textit{``how does the liver work?''} but rejects
a presupposition of \textit{``what's the best liver detox cleanse?''}.
Previous computational research has studied discourse acts in
conversational speech~\cite{stolcke-etal-2000-dialogue}, scientific
articles~\cite{teufel2002summarizing}, and on
Reddit~\cite{zhang2017characterizing}, but these acts are too general
(e.g. \textit{statement}, \textit{question}, \textit{announcement})
making them ill-fitting for studying QA.
Others have characterized the discourse structure of \textit{general}
human and machine-generated text~\cite{kim-etal-2024-threads} using \textit{rhetorical structure theory}~\cite[RST]{mann1987rhetorical}, a theoretical linguistic framework to describe text coherence with relations between hierarchically organized text segments.

Perhaps most relevant, \citet{xu2022we} study the discourse structure of answers from the ELI5, WebGPT~\cite{nakano2021webgpt}, and Natural Questions~\cite{kwiatkowski-etal-2019-natural} datasets, presenting six roles of answer sentences (\textit{summary}, \textit{answer}, \textit{example}, \textit{auxiliary info}, \textit{miscellaneous}, and \textit{organizational}), but these roles focus more on how answers \textit{organize} information, whereas
our goal is to characterize what the answerer is \emph{doing} in relation
to the question asked---including moves like seeking
clarification~\cite{ICLR2025_97e2df4b} or challenging a question's
premise~\cite{yu-etal-2023-crepe}---especially across communities beyond \subreddit{r/explainlikeimfive}.
Additionally, they discount questions that are not
``interpretable'', have false presuppositions, or contain
subquestions, but given that humans often ask these types of questions
to LLMs~\cite{kim2021linguist, min-etal-2020-ambigqa}, we choose to
model how answerers respond to these questions as a part of
their answering strategy instead of filtering them out.

\paragraph{Question Interpretation.} When responding to a question $Q$, answerers may address one or more readings, or \textit{interpretations}, of $Q$. 
This may be because $Q$ is ambiguous~\cite{min-etal-2020-ambigqa} or
vague~\cite{stengel2021human} or because overanswering serves an
underlying goal~\cite[e.g., to teach the asker
something,][]{tsvilodub2023overinformative}.
As such, we also study how answerers address \textit{interpretations} of questions when crafting a response.
Modeling what answerers address is related to the Questions Under Discussion framework~\cite[QUD, ][]{roberts2012information}, which proposes that discourse is organized around questions that sentences or utterances may implicitly or explicitly resolve.
Computational work on QUDs~\cite{ko2023discourse, wu2023qudeval, namuduriqudsim} and ``question-focused'' discourse~\cite{wang-etal-2025-uncovering} recovers the implicit questions that drive \textit{discourse coherence} rather than modeling how answerers select and respond to readings of an explicit question.
\dt{} pairs discourse acts in an answer with an interpretation of the original question to not only surface \textit{how} the answerer responds, but also specifically \textit{what} they responded to.

\section{\dt: Structured Discourse Representations of Answers}
\label{sec:discotrace}
Here, we describe constructing the \dt{} of an answer, revealing the high-level operations an answerer uses to address a question's information need.
Our pipeline ingests a question $Q$ and an answer $A$, and outputs a \textit{strategy} \strategy---a \textit{sequence} consisting of discourse act-interpretation tuples $\{(a_1, i_1), (a_2, i_2)...(a_n, i_n)\}$ (see Fig.~\ref{fig:teaser-example}), in three stages: (1) rhetorical structure theory (RST)-based parsing (\S\ref{subsec:rst}), (2) discourse act labeling (\S\ref{subsec:actions}), (3) interpretation labeling on eligible discourse acts (\S\ref{subsec:interps}).

\begin{figure*}
\centering
\includegraphics[scale=0.6]{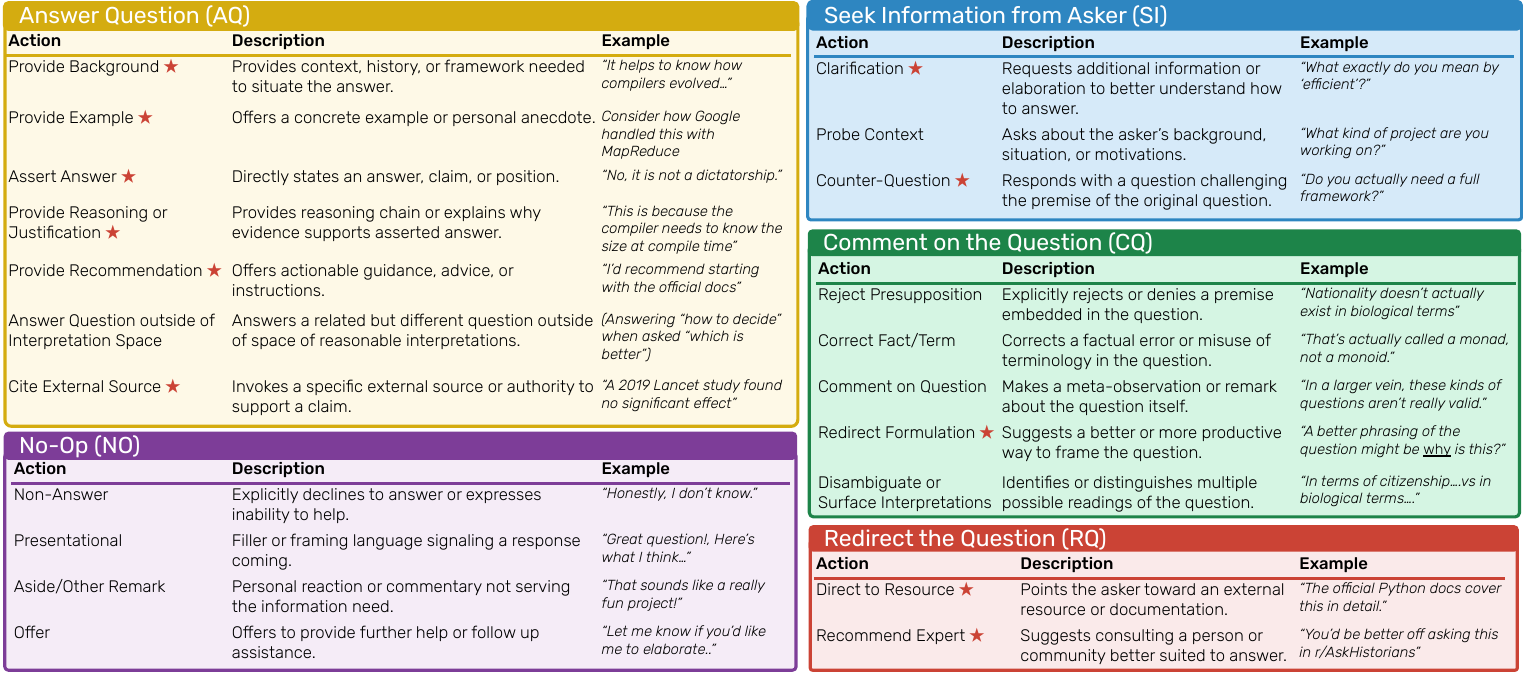}
\caption{Each segment in a \dt{} is labeled with one of these actions, grouped into five broad families. The full \dt{} pairs each of these with a question-specific interpretation.}
\label{fig:action-ontology}
\end{figure*}

\subsection{Step 1: Preprocessing Answers with RST}
\label{subsec:rst}

Rhetorical structure theory (RST) explains the coherence and
structure of text by identifying relationships between textual
subsegments~\cite{mann1987rhetorical}.
RST represents a text $t$ as a binary constituent tree (Fig.~\ref{fig:rst-parsing}) with
clause-like leaves representing atomic spans of $t$ (elementary
discourse units, or EDUs) that recursively merge until the
root represents the full text.

We preprocess an input answer $A$ using a competitive end-to-end RST parser~\cite{chistova-2024-bilingual}\footnote{This lightweight model (\url{https://huggingface.co/tchewik/isanlp_rst_v3}) pre-trained on the RST Discourse Treebank~\cite{carlson-etal-2001-building} earns a span micro F1 score of 78.7 as compared to heavier-weight \texttt{llama-2-70b} earning only 1 point more at 79.8~\cite{maekawa-etal-2024-obtain}.}
based on \model{xlm-roberta-large} to construct its RST tree using 18 coarse-grained rhetorical relations (Appendix~\ref{appendix:rst}).
Preprocessing with RST provides a useful discourse scaffolding, but it alone is insufficient since: (1) RST relations hold \textit{between} textual segments rather than between segments and some external question, and (2) EDUs are too short to serve as segments themselves to determine a higher-level discourse relation with the question (see leaf nodes, which are EDUs,  in Fig.~\ref{fig:rst-parsing}).
Labeling segments with discourse acts (\S\ref{subsec:actions}) addresses the first concern.
For the second, we construct coarser \textbf{action segments} with one more preprocessing step.

Motivated by the observation that some relations are more likely to signal shifts than others (e.g. in subtopics ~\cite{cardoso2017subtopic}), we identify a subset of RST relations which we call \textit{boundary relations} (\textit{contrast}, \textit{comparison}, \textit{topic-change}, \textit{evaluation}, \textit{summary}, and \textit{background}) where we expect a rhetorical shift between children; children with these boundary relations likely correspond to different discourse acts.
Since a relation may signal a shift under some nuclearity
configurations but not others, we define boundary relations as
relation--nuclearity pairs (Table~\ref{tab:boundary-relations},
Appendix).

We recursively traverse the RST tree, splitting a node's children into separate segments when the node's relation is a boundary relation, and preserving any splits that arise from deeper in the tree. Unsplit subtrees are collapsed into single spans. 
See appendix for the full procedure (Algorithm~\ref{alg:segment}).
This yields text spans (or action \textit{segments})
with \textit{suspected} distinct discourse acts which we label with
our question-based act ontology next.

\subsection{Step 2: Question-Related Discourse Acts}
\label{subsec:actions}

While prior work has studied specific phenomena in isolation (e.g., rejecting false presuppositions~\cite{yu-etal-2023-crepe}, follow-up questions~\cite{meng-etal-2023-followupqg}, or clarifications~\cite{majumder-etal-2021-ask}), we develop a general ontology of 21 discourse action types designed to capture the full breadth of rhetorical moves answerers may employ across diverse question-answering settings.

\paragraph{Action Ontology.} Our ontology of 21 discourse acts (Figure~\ref{fig:action-ontology}) is divided into five families. Answerers can \textit{\underline{a}nswer} the \underline{q}uestion (AQ), \textit{\underline{c}omment} on the \underline{q}uestion (CQ), \textit{\underline{s}eek} more \underline{i}nformation from the user (SI), \textit{\underline{r}edirect} the \underline{q}uestion to a more appropriate source (RQ), or \textit{no-op} (NO) by writing text that does not substantively advance the answer.
We discuss our iterative, data-driven approach to ontology construction in Appendix~\ref{appendix:ontology-construction}.

\paragraph{Parsing.} 
Answers can, and often do, employ multiple discourse acts from our ontology (Fig.~\ref{fig:rst-parsing}).
We label each action segment from \S\ref{subsec:rst} with a discourse act (or, when no act is appropriate, \texttt{NONE}\footnote{The automatic labeler's usage of the \texttt{NONE} label remains under 4\% of all segments in our dataset of Reddit questions, showing the broad coverage of our ontology (Table~\ref{tab:none_rates}).}) using \model{gpt-4.1-2025-04-14} (Prompt~\ref{prompt:discourse-act-tagging}).
Since RST-based action segments are the result of a coarse preprocessing step, they \textit{may} still contain multiple acts that must be broken down further.
To handle this, we provide each EDU within each action segment as text subspans to the model, which can label the segment as a whole or assign different labels to individual EDUs if they serve different discourse functions (Appendix~\ref{appendix:act-labeling}).
To reduce ambiguity at the boundaries between discourse acts, we also provide the model with the preceding segment and its predicted label as context: it can predict the \textit{continuation} of an act or the \textit{initiation} of a new act, as is standard in sequence tagging~\cite{ramshaw-marcus-1995-text}.
This step produces a sequence of discourse act tags corresponding to subsegments. 

\paragraph{Human Validation.} Two external annotators label 200 segments (arising from 55 answers uniformly randomly sampled across the nine subreddits) using our action ontology, as well as have them match eligible actions to interpretations in our generated space.
Appendix~\ref{appendix:guidelines} includes the guidelines we show to annotators.
The discourse act agreement of the two external annotators is $\kappa=0.75$ at the family level (six-way labeling: five families +  \model{NONE}) as measured with Cohen's Kappa, and $\kappa=0.62$ at the individual action level, indicating substantial agreement~\cite{artstein-poesio-2008-survey}.
Family-level agreement between each annotator $A_1/A_2$ and \model{gpt-4.1} is $\kappa=0.74/0.73$ and act-level is $\kappa=0.66/0.52$.
Inter-annotator disagreement mostly concentrates within-family: of all segment disagreements, 71.9\% are within-family, compared to 28.1\% across family.\footnote{The top three within-family confusion pairs were: \abr{Assert Answer}$\Leftrightarrow$\abr{Provide Reasoning} (AQ), \abr{Assert Answer}$\Leftrightarrow$\abr{Provide Example} (AQ), and \abr{Provide Background}$\Leftrightarrow$\abr{Provide Reasoning} (AQ). The top three cross-family confusion pairs were \abr{Cite External Source}$\Leftrightarrow$\abr{Direct to Resource} (AQ vs. RQ), \abr{Assert Answer}$\Leftrightarrow$\abr{Reject Presupposition} (AQ vs. CQ), and \abr{Assert Answer}$\Leftrightarrow$\abr{Aside/Other Remark} (AQ vs. NO).}

\subsection{Step 3: Pairing Acts with Interpretations}
\label{subsec:interps}

The rhetorical strategy \strategy~produced by \dt~must not only describe \textit{how} an answerer chooses to respond, but also to \textit{which} interpretation(s) of the question.
A question~$Q$ may admit multiple valid readings or interpretations~\cite{min-etal-2020-ambigqa}.
This may be for a number of reasons: $Q$ was not fully formed in an asker's mind~\cite{taylor1962process}, the asker is not an expert in the topic or domain of $Q$, or $Q$ is inherently open-ended.
Understanding which \textit{interpretation(s)} of $Q$ an answerer addresses is therefore crucial to characterizing their strategy.
For example, the question in Fig.~\ref{fig:teaser-example} may reasonably admit at least three interpretations, and answerers may choose to adopt one or more of these interpretations, ignore others, or even surface the ambiguity explicitly.

\paragraph{Act Eligibility.} Not all discourse acts assume or address a particular interpretation (presupposition rejection cannot adopt an interpretation, since it contests the question's premise rather than attempting to answer it), while others, like clarification, may adopt an interpretation, since they often seek to resolve ambiguity with respect to a specific reading the answerer has in mind.
Our ontology marks which acts are eligible to be paired with an interpretation (marked \annotateinterp{} in Fig.~\ref{fig:action-ontology}), meaning that the act reasonably addresses that interpretation.

\paragraph{Interpretation Generation and Pairing with Discourse Acts.}
We first establish a set of reasonable interpretations of a question $Q$.
Prior work identifies interpretations of $Q$ via crowdsourcing~\cite{min-etal-2020-ambigqa} or with retrieval-augmented generation~\cite{kim-etal-2023-tree} or direct prompting~\cite{srikanth2025no}.
\citet{namuduriqudsim} use direct prompting with \model{gpt-4o} to generate questions under discussion.
We generate a space of reasonable, distinct interpretations $S_I$ from $Q$ via direct prompting: first, we provide $Q$ and some contextual information $C$\footnote{In our experiments, $C$ is a short description of the subreddit in which the question was asked (Table~\ref{tab:subreddits}), but can be adapted to any other setting simply by describing the context in which $Q$ was asked.} to two LLMs (\model{gpt-4.1-2025-04-14} and \model{claude-sonnet-4.6}) and prompt them to generate a variable number of generations using Prompt~\ref{prompt:interpretation-generation}.
Models may abstain from generating any interpretations if they deem the
question clear.
We pool generated interpretations from both models and use the deduplication algorithm from \citet{srikanth2025no} to
construct $S_I$ (Table~\ref{tab:interp_stats} shows $|S_I|$ for different subreddits).\footnote{We use \model{Qwen3-Embedding-8B} as the
instruction-tuned embedder instead. All other hyperparameters are the same.}
Finally, using the act label predictions from \S\ref{subsec:actions}, \model{gpt-4.1-2025-04-14} assigns an interpretation from $S_I$ to each eligible act (Prompt~\ref{prompt:interpretation-labeling}).

\paragraph{Human Validation.} The same two annotators from \S\ref{subsec:actions} assign interpretations from $S_I$ to their eligible acts or abstain if they believe the act does not adopt one for the 200 segments.
On whether a segment addressed an interpretation at all (binary task): annotators achieve a Cohen's $k=0.71$.
For segments where \textit{both} annotators assign an interpretation, they agree on the exact interpretation 74\% of the time.
Annotator-model agreement on the binary task for $A_1 / A_2$ is $k=0.78/0.51$ with exact interpretation agreement of $74\%/73\%$ when both the annotator and model assign one.

\section{Discourse Across Humans and LLMs}
\label{sec:communities}
\dt{} can describe answers, compare answers for preference datasets, or help generate human discourse-aligned answers.
Here, \dt{} surfaces answering strategy differences (1) across QA communities (\S\ref{subsec:human-communities}), signaling that the \textit{objectives} that answerers optimize across communities are distinct, and (2) across humans and LLMs (\S\ref{subsec:llm-answering-discourse}), suggesting that by default, LLMs model a generic, narrow answering strategy. 
Even when prompted to mimic a particular human community, LLMs fail to capture its full strategy diversity.

\subsection{Reddit Communities as Answer Data}
\label{subsec:data}
We source answers from Reddit\footnote{Obtained from the \texttt{pushshift} dump
(\url{https://github.com/Watchful1/PushshiftDumps}).}, since it hosts communities (\textit{subreddits}) with different answering objectives.
Our nine selected communities (Table~\ref{tab:subreddits}) vary along several axes: expert forums (\subreddit{AskEconomics}, \subreddit{AskHistorians}, \subreddit{asklinguistics}), general QA (\subreddit{NoStupidQuestions}, \subreddit{OutOfTheLoop}, \subreddit{explainlikeimfive}), experiential communities (\subreddit{ScienceBasedParenting}, \subreddit{beyondthebump}), and the topic-based general \subreddit{history}.\footnote{We include \subreddit{history} to compare with \subreddit{AskHistorians} as it shares subject matter, but with less expertise asymmetry and no dedicated QA format.}
For each community, we randomly sample 300 questions and their surviving top-level comments from a pool of posts filtered by quality heuristics (Appendix~\ref{appendix:question-selection}).
We then generate a \dt~for each of the 18,968 top-level comments (answers) to the 2,700 questions in our dataset.

\begin{figure*}[t!]
\centering
\begin{subfigure}[t]{0.46\linewidth}
    \centering
    \includegraphics[width=\linewidth]{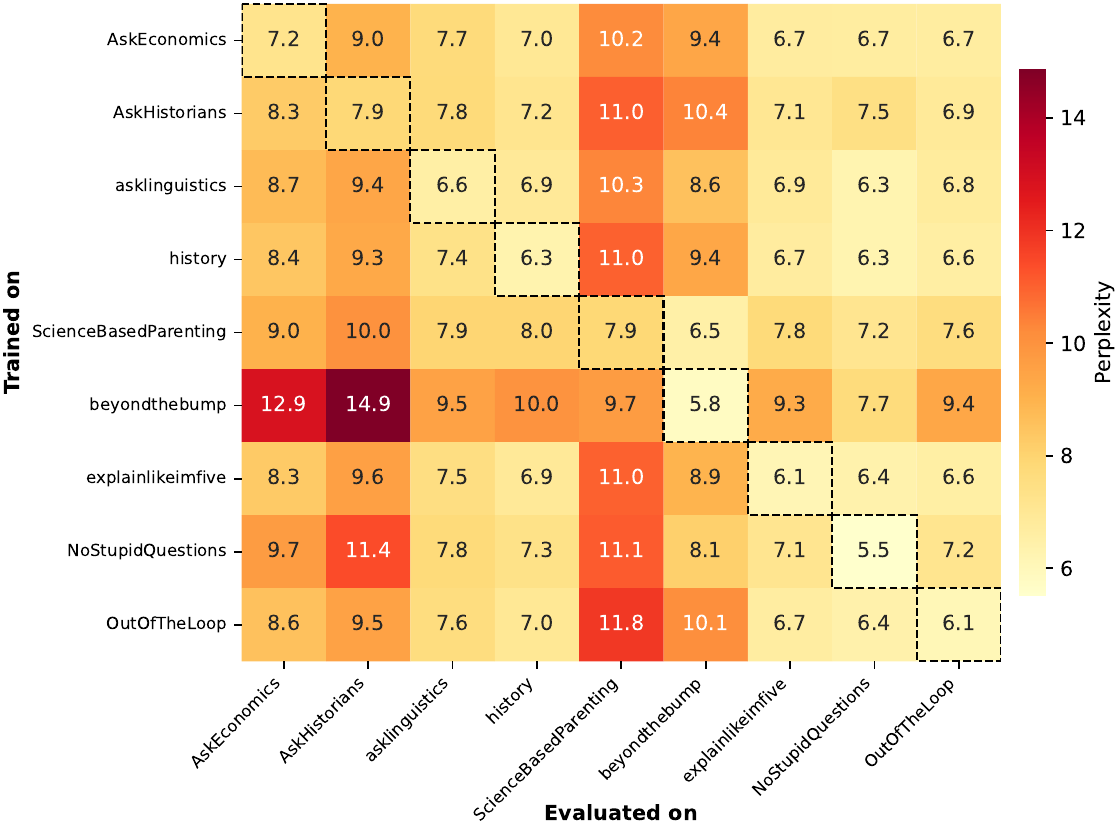}
    \caption{Cross-perplexity of the discourse act bigram model
\emph{across subreddits}: trained and evaluated on human answers.}
    \label{fig:human-v-human}
\end{subfigure}
\hfill
\begin{subfigure}[t]{0.46\linewidth}
    \centering
    \includegraphics[width=\linewidth]{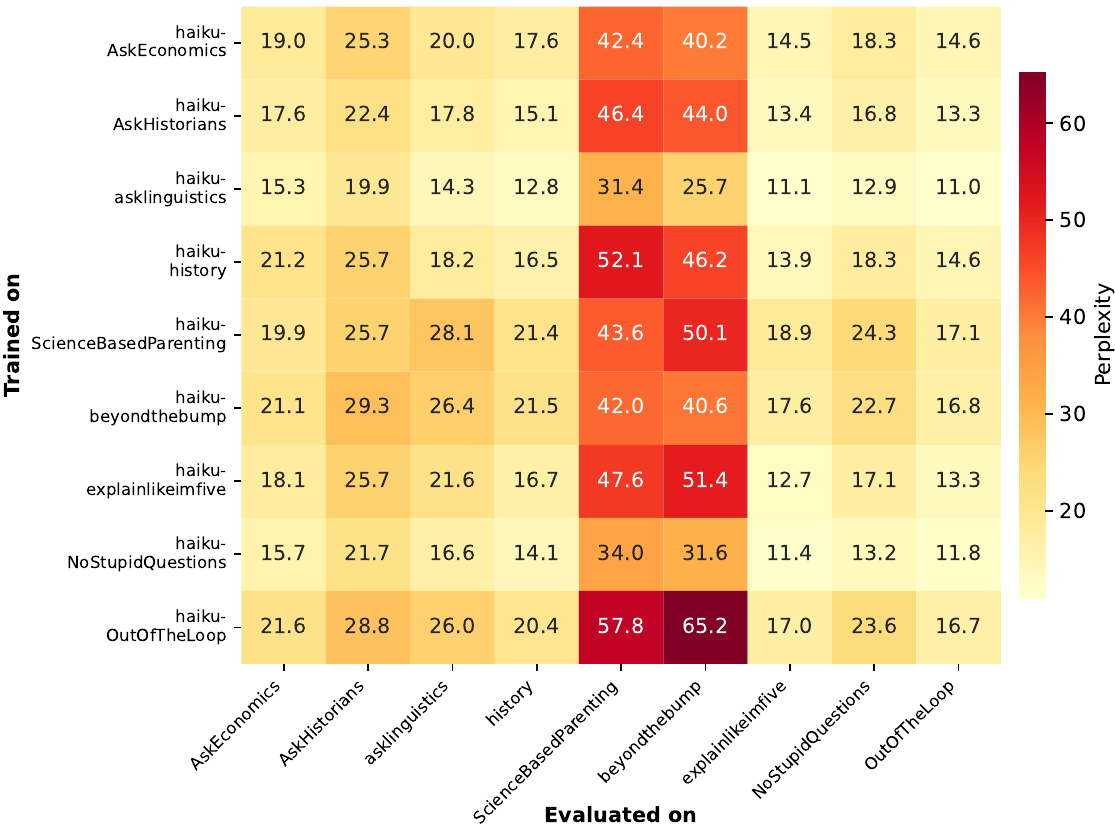}
    \caption{Cross-perplexity of the discourse act bigram model trained on
\model{claude-haiku-4.5} answers, evaluated on human answers.}
    \label{fig:haiku-v-human}
\end{subfigure}
\label{fig:comm-perplexity}
\caption{\dt{} helps compare how different answering \emph{strategies} are across different subreddits. We train bigram models on \textit{act} sequences from one community $C_\text{train}$ (rows) and evaluate on another $C_\text{eval}$ (columns). \textbf{(a)} Bigram models are trained on human answers; the diagonal is each community's self-perplexity. \textbf{(b)} Bigram models are trained on \model{claude-haiku-4.5} answers to questions from $C_\text{train}$; here the diagonal compares the LLM's answers against human answers \emph{in the same community}. \textbf{Note the differing scales: even the best LLM--human values (b) exceed most human--human values (a).}} 
\end{figure*}

\subsection{Discourse Across \textit{Human} Communities}
\label{subsec:human-communities}

\paragraph{Setup.} We use the \dt{} of each answer across all nine subreddits in our dataset (\S\ref{subsec:data}) to analyze distributions of answering strategies across communities.
We train a bigram model over discourse act sequences for each community $C$, estimating $p(a_\text{next} \g  a_\text{prev})$ to capture both the acts and their ordering as an estimation of $C$'s preferred answering ``policies.''
We include start and end tokens to model how answers open and close.
We compare community policies by computing a cross-perplexity matrix over all pairs of subreddits $(C_1, C_2)$ by taking a trained bigram model on all answers in $C_1$ and evaluating it on answers in $C_2$.
Perplexity here reflects how surprising one community's discourse act patterns are under another community's model---high perplexity indicates divergent answering policies.
This cross-perplexity matrix reveals:

\paragraph{Expert and generic QA communities differ sharply in answering strategies.} Domain-\textit{agnostic} QA communities like \subreddit{NoStupidQuestions} or \subreddit{explainlikeimfive} are predicted well by virtually all other communities (Fig.~\ref{fig:human-v-human}, cols. 7--9), suggesting a generic baseline answering style.
In contrast, generic communities (rows/cols. 7-8) are poor predictors of expert communities (rows/cols. 1–3), while expert communities predict each other relatively well (\subreddit{AskEconomics} $\rightarrow$ \subreddit{asklinguistics}, \subreddit{AskHistorians} $\rightarrow$ \subreddit{AskEconomics}), suggesting a distinct expert answering style that may generalize across disciplines.
Answerers in \subreddit{AskHistorians} are more likely to ask clarification questions, cite external sources, and provide background context, while answerers in \subreddit{explainlikeimfive} take a practical approach, directly asserting an answer as well as providing recommendations or advice to the asker (Appendix, Fig.~\ref{fig:act-comparison-eli5-v-askhistorians}).
This expert-generic distinction is critical, as most LFQA research analyzes the ELI5 dataset, which reflects only a small slice of the diverse range of human answering policies that may satisfy different user preferences.

\paragraph{Shared topic does not imply shared rhetorical strategy.} 
The communities \subreddit{history} and \subreddit{AskHistorians} both discuss history, but are poor predictors of each other's discourse structure.
\subreddit{AskHistorians}, with its strict moderation requiring sourced and comprehensive answers, develops a more constrained discourse style that can partially account for \subreddit{history}'s looser patterns, but not vice versa.
Similarly, \subreddit{ScienceBasedParenting} and \subreddit{beyondthebump} both revolve around parenting young children.
Answering strategies in the latter are unsurprising to a model trained on the former, but the reverse is not true.
While both of these communities emphasize experience sharing and community seeking when answering, \subreddit{ScienceBasedParenting} additionally mandates the citation of peer-reviewed articles to support claims, resulting in a broader range of answering strategies.\footnote{\tiny\url{https://www.reddit.com/r/ScienceBasedParenting/comments/1f9vwjt/welcome_and_introduction_september_2024_update/}}

These patterns, recoverable from discourse act sequences alone, illustrate that QA \textbf{is not one size fits all}: similar questions posed in different communities may yield vastly different answering strategies. 
Where prior work conditions on subreddit name during preference optimization~\cite{kumar-etal-2025-compo}, conflating topic, answering strategy, and community norms in a single token, \dt{} offers an interpretable and generalizable alternative, surfacing explicit answering objectives (e.g. \textit{contextualize} when answering) that future work may use when personalizing answer generation.

\subsection{Human vs. LLM Answering}
\label{subsec:llm-answering-discourse}
LLMs are often post-trained on preference data that collapses distinctive preferences~\cite{bakker2022fine}.
We ask two questions: whether LLMs' default answering strategies exhibit
the cross-community rhetorical diversity we observe in humans
(\S\ref{subsec:human-communities}), and whether prompting with community
context recovers the strategies of a particular community.
\begin{figure}[t!]
\centering
\includegraphics[scale=0.7]{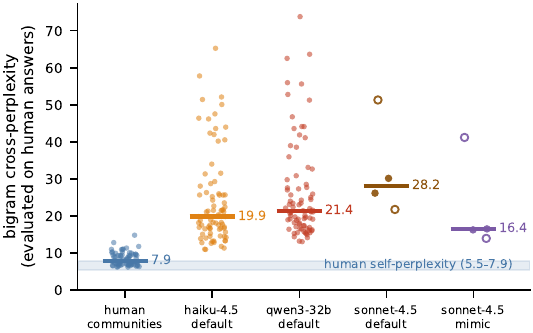}
\caption{LLM answering strategies diverge from every human community (\S\ref{subsec:llm-answering-discourse}).
Each point is the perplexity of a bigram model over discourse act
sequences from one source, evaluated on human answers from one community
(72 distinct human community pairs, 81 pairs per default LLM, and 4 per
\model{sonnet-4.5} condition where 2 open points are cross-community and 2 filled are within-community). Lines are medians; the band spans human
within-community self-perplexity. All LLM settings diverge from human
communities more than communities do from one another; community
guidelines (``mimic'') narrow but do not close the gap.}
\label{fig:gap}
\end{figure}
\paragraph{Default Behavior.}
We prompt both a closed (\model{claude-haiku-4.5}) and open source (\model{qwen3-32b}) LLM to answer questions from all communities without providing any context or instruction beyond the question itself, restricting responses to 1K tokens.
Then, we train bigram models over LLM answer action sequences (\S\ref{subsec:human-communities}).
Cross-perplexity between LLM-generated answers and human answers is much higher than between any human communities (Fig.~\ref{fig:gap}), showing that LLM discourse structures diverge significantly from human communities.
Fig.~\ref{fig:haiku-v-human} shows the full per-community
cross-perplexity matrix for \model{haiku} (Fig.~\ref{fig:qwen-v-human},
Appendix, for \model{qwen}): both models best predict human strategies
in generic, explanation-oriented communities
(\subreddit{explainlikeimfive}, \subreddit{OutOfTheLoop}).
Measuring perplexity between LLM answers across different subreddits (Figs.~\ref{fig:haiku-v-haiku} and \ref{fig:qwen-v-qwen} in appendix) reveals that they deploy a largely uniform discourse structure \textit{across question types and topics}.

\paragraph{Adapting to a Community.}
\begin{figure}[t!]
\centering
\includegraphics[scale=0.55]{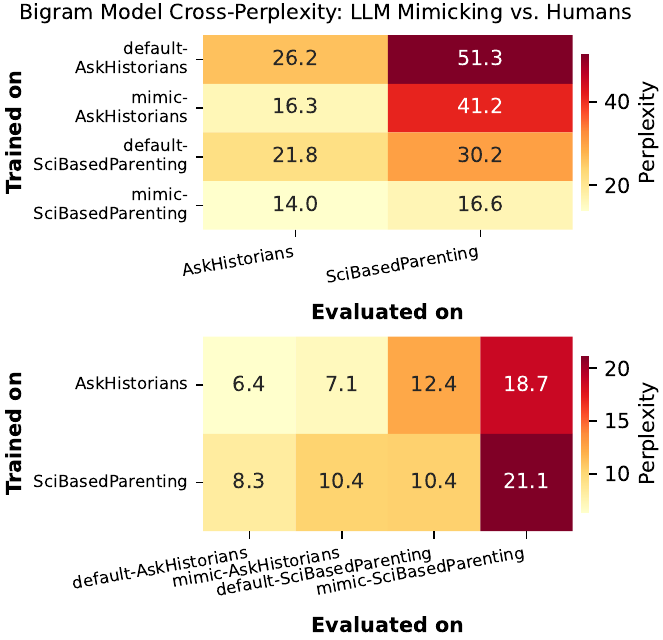}
\caption{Prompting \model{claude-sonnet-4.5} with the community guidelines to answer questions as a member of two subreddits \subreddit{AskHistorians} and \subreddit{ScienceBasedParenting} (\model{mimic}) is insufficient to match the norms of the underlying community. Human answers are poorly predicted by bigram models trained on LLM answers (top), while LLM answers are far more predictable under human bigram models (bottom). This asymmetry indicates that LLMs deploy a narrower range of strategies, one that falls largely within (but does not cover) the human strategy distribution.}
\label{fig:mimic}
\end{figure}

In the default setting, LLMs are unaware of the community addressed by a question.
While analyzing the default discourse structures of LLMs helps us model a straightforward user-LLM interaction without any intervention, it does not necessarily reveal whether LLMs have the capacity to mimic a certain community's answering ``policy'' when prompted to do so. 
To measure whether this absence of context inhibits rhetorical diversity in LLM answers, we provide a more powerful closed-source model (\model{claude-sonnet-4.5}) the community guidelines of two of the most distinctive communities in our dataset, \subreddit{AskHistorians}\footnote{\tiny\url{https://www.reddit.com/r/AskHistorians/wiki/rules/\#answers}} and \subreddit{ScienceBasedParenting}, and ask it to answer as if it were a subreddit member (Prompt~\ref{prompt:mimic}).
We also prompt it in the default setting for direct comparison.
Mimicking \subreddit{AskHistorians} this way drops perplexity from 26.2 to 16.3, but human self-perplexity in this community is 7.9 (similar in \subreddit{SciBasedParenting}, Fig.~\ref{fig:human-v-human}): a sizable gap remains even after the LLM is given explicit, detailed instructions on how to answer.
Notably, these guidelines already prescribe behavior at the discourse level (avoid hedging and speculation, suppress fillers, write complete rather than partial answers), yet the \textbf{model does not translate these behavioral specifications into the rhetorical patterns of community answers.}
This suggests that community rhetorical norms run deeper than what answering instructions can encode: the lack of rhetorical diversity is not merely an issue of context availability, but may reflect a limitation in LLMs' pragmatic reasoning about user information needs~\cite{taylor1962process} when constructing an answer.

\section{Interpretation Selection}
\label{sec:interpretations}
\begin{figure}[t!]
\centering
\includegraphics[scale=0.54]{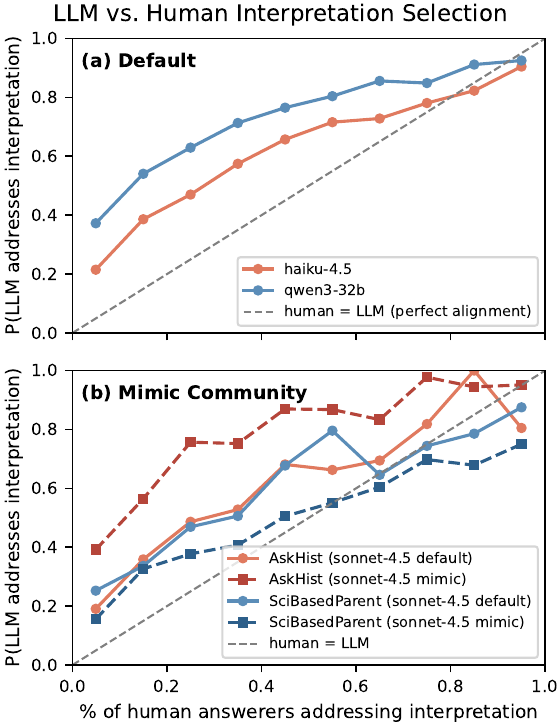}
\caption{LLMs address low probability interpretations that humans do not address (top). This behavior can get \textit{worse} when it mimics answerers in a specific community (bottom, \subreddit{AskHist} default to mimic).}
\label{fig:llm-selection}
\end{figure}

Answering a question involves pragmatic reasoning about the asker's intent~\cite{grice1975logic} and \textit{information need} to provide maximally relevant information~\cite{sperber1986relevance}. 
\dt{} not only enables the analysis of discourse \textit{acts} as in \S\ref{sec:communities}, it also reveals the \textit{content} choices of answerers by matching \textit{interpretations} of the question with discourse acts.
This lets us compute metrics such as interpretation overlap between two answers to the same question, the proportion of an answer dedicated to an interpretation, and the discourse act type used to address it. 
Our analyses reveal that LLMs deploy narrower, uniform answering policies than humans, but do they align in their choices of \textit{interpretation}?

\paragraph{Humans.} Humans in some communities address more of the interpretation space $S_I$ of a question than others (Table \ref{tab:interp_stats}): \subreddit{explainlikeimfive} and \subreddit{AskHistorians} answers cover 40-45\% of $S_I$, versus only 34\% in \subreddit{NoStupidQuestions}.
This illustrates that the proportion of interpretations addressed in an answer can be influenced by the objectives and preferences of a community. %
\begin{table}
\centering
\resizebox{\columnwidth}{!}{
\begin{tabular}{clcc} 
\toprule
\textbf{prompt mode} & \textbf{answerer} & \textbf{mean} & \textbf{median} \\ 
\cmidrule{1-1}\cmidrule{2-4}
--- & human & 0.62 & 0.53 \\ 
\midrule
\multirow{2}{*}{default} & \texttt{haiku-4.5} & 0.46 & 0.40 \\
 & \texttt{qwen3-32b} & 0.40 & 0.31 \\ 
\midrule
default & \texttt{claude-sonnet-4.5}* & 0.48 & 0.40 \\
mimic & \texttt{claude-sonnet-4.5}* & 0.41 & 0.33 \\
\bottomrule
\end{tabular}}
\caption{Mean/median proportion of eligible segments dedicated to each interpretation. Humans dedicate more segments per interpretation, producing more focused answers. * indicates the model was evaluated only on \model{AskHistorians} and \model{ScienceBasedParenting} (the two ``mimic'' communities).}
\label{tab:interpretation-dedication-short}
\end{table}

\paragraph{LLMs.} We study interpretation selection of the two LLMs from \S\ref{subsec:llm-answering-discourse}.
For each interpretation $i \in S_I$ across questions from all communities, we (1) compute the proportion of human answers addressing $i$, (2) bin interpretations by this frequency, and (3) plot the average probability of an LLM addressing $i$ per bin (Fig.~\ref{fig:llm-selection}a).
The diagonal indicates perfect alignment with human interpretation preferences.
Placement above it indicates overanswering: addressing interpretations that humans rarely address.
With default prompting, both LLMs systematically overaddress: even when fewer than 10\% of humans address $i$, \model{haiku} addresses it over 20\% of the time and \model{qwen} 40\%.
Instead of reasoning about exact asker information need, LLMs opt to hedge and deluge the user with information.

Using the same \dt{}s from \model{sonnet-4.5} mimicking community members (\S\ref{subsec:llm-answering-discourse}) shows that mimicking does not help, and can even worsen the pattern (Fig.~\ref{fig:llm-selection}b, \subreddit{AskHist}).
Computing the average proportion of eligible discourse acts in an answer devoted to an interpretation $i$ reveals that humans favor depth over breadth (Table~\ref{tab:interpretation-dedication-short}), opting to address fewer interpretations more exhaustively.
This pattern persists across communities with different rhetorical strategies, suggesting it is not simply a stylistic preference, but may be a core feature of human pragmatic reasoning when answering a question.
LLM-generated answers are \textit{systematically lengthier} (Table~\ref{tab:interpretation-dedication-long}), corroborating prior work~\cite{poddar2025brevity}, and additional length is the cost of addressing interpretations that the asker likely didn’t intend.
Future work must strengthen LLM pragmatic reasoning ability in QA to, like humans, hone in on likely interpretations that matter.

\section{Conclusion}
\label{sec:conclusion}
\textbf{QA is not one size fits all}. \dt{} reveals that communities
take different strategies to answer questions, driven by different
preferences, needs, and norms, and that LLMs do not answer with the full
rhetorical breadth of humans.
More broadly, \dt{} turns rhetorical strategy from an implicit property
of answers into an explicit representation that systems can use to plan.
It can characterize how humans respond to ambiguity (surfacing it versus
committing to an interpretation).
It can also support preference-sensitive answering: an LLM could
generate a \dt{} as a high-level plan conditioned on a user's needs,
then populate content accordingly.
For example, a consumer health assistant~\cite{srikanth-etal-2024-pregnant} that
maximizes simplicity and minimizes harm might prioritize \abr{Assert
Answer} with \abr{Provide Reasoning} and routinely \abr{Recommend
Expert}; the same planner serving clinicians would instead favor
\abr{Cite External Source} and \abr{Surface Interpretations}, with less
\abr{Provide Background}, since expertise can be assumed.
\dt{} can also disentangle preference judgments that may conflate strategy
with execution~\cite{hu-etal-2023-decipherpref}: fixing a target \dt{}
across candidate answers isolates execution quality (clarity, accuracy,
prose), while varying the plan isolates preferences over strategy
itself.
Finally, \dt{} could also be used to study internal model representations of
answering strategies.

\section*{Acknowledgments}
We thank the anonymous reviewers as well as the members of the University of Maryland CLIP lab for their thoughtful feedback.
We also thank our annotators: expert linguists from Upwork, as well as Maya Srikanth and Rupak Sarkar.
This work was supported by NIH Award No. R01MD016037 (Srikanth and Boyd-Graber) and NSF CAREER Award No. 2339746 (Rudinger).
Boyd-Graber is also supported by a gift from Adobe Corporation.
The content is solely the responsibility of the authors and does not necessarily represent the official views of the sponsors. 
The funders had no role in study design, data collection and analysis, decision to publish, or preparation of the paper.

\section*{Limitations}
While this work studies answers across a variety of online communities, questions and answers are restricted to those from Reddit: we do not include QA pairs from other platforms such as Quora or StackOverflow.
Though different subreddits have different answering strategies, these may be affected by the platform as a whole, and are not necessarily representative of humans across all contexts and platforms. 
All questions and answers in our study are in English, and our
discourse act ontology was developed on English Reddit data.
Whether the ontology and our findings transfer to other languages or answerers from diverse cultural and linguistic backgrounds and platforms is an open question.

Our experiments do not use the body of Reddit posts as extra context, similar to other datasets such as ELI5~\cite{fan2019eli5} and Stanford Human Preferences 2 (SHP-2)~\cite{pmlr-v162-ethayarajh22a}.
Beyond precedent, we opt to do this to ensure a fair comparison across questions, since some posts contain no body text and others have several paragraphs.
We also study answers from top-level comments only, ignoring follow-up replies or multi-turn exchanges where the original asker may have refined their question or information need, possibly influencing the strategies of other answerers on the thread.

Because we filter answers by upvotes (Appendix~\ref{appendix:question-selection}), the human answers we study are those a particular community \textit{rewarded}, not the full range answerers \textit{attempted}. 
Our characterization of human strategies is therefore best
read as a characterization of \emph{community-endorsed} strategies.
This aligns with our aim of modeling community answering norms, but it
means low-scoring strategies (which may include reasonable moves a
community simply disfavors) are underrepresented, and that our
human--LLM comparisons measure distance from \textit{endorsed} norms rather
than from \textit{all} human answering behavior.

Lastly, our community-mimicking experiment covers two communities, and a full
cross-community grid would more comprehensively test whether context
can recover human strategy diversity.

\bibliography{custom}

\begin{thebibliography}{55}
\expandafter\ifx\csname natexlab\endcsname\relax\def\natexlab#1{#1}\fi

\bibitem[{Adamic et~al.(2008)Adamic, Zhang, Bakshy, and Ackerman}]{adamic2008knowledge}
Lada~A Adamic, Jun Zhang, Eytan Bakshy, and Mark~S Ackerman. 2008.
\newblock Knowledge sharing and yahoo answers: everyone knows something.
\newblock In \emph{Proceedings of the 17th international conference on World Wide Web}, pages 665--674.

\bibitem[{Artstein and Poesio(2008)}]{artstein-poesio-2008-survey}
Ron Artstein and Massimo Poesio. 2008.
\newblock \href {https://doi.org/10.1162/coli.07-034-R2} {Survey article: Inter-coder agreement for computational linguistics}.
\newblock \emph{Computational Linguistics}, 34(4):555--596.

\bibitem[{Bakker et~al.(2022)Bakker, Chadwick, Sheahan, Tessler, Campbell-Gillingham, Balaguer, McAleese, Glaese, Aslanides, Botvinick et~al.}]{bakker2022fine}
Michiel Bakker, Martin Chadwick, Hannah Sheahan, Michael Tessler, Lucy Campbell-Gillingham, Jan Balaguer, Nat McAleese, Amelia Glaese, John Aslanides, Matt Botvinick, et~al. 2022.
\newblock Fine-tuning language models to find agreement among humans with diverse preferences.
\newblock \emph{Advances in neural information processing systems}, 35:38176--38189.

\bibitem[{Cao and Wang(2021)}]{cao-wang-2021-controllable}
Shuyang Cao and Lu~Wang. 2021.
\newblock \href {https://doi.org/10.18653/v1/2021.acl-long.502} {Controllable open-ended question generation with a new question type ontology}.
\newblock In \emph{Proceedings of the 59th Annual Meeting of the Association for Computational Linguistics and the 11th International Joint Conference on Natural Language Processing (Volume 1: Long Papers)}, pages 6424--6439, Online. Association for Computational Linguistics.

\bibitem[{Cardoso et~al.(2017)Cardoso, Pardo, and Taboada}]{cardoso2017subtopic}
Paula~CF Cardoso, Thiago~AS Pardo, and Maite Taboada. 2017.
\newblock Subtopic annotation and automatic segmentation for news texts in brazilian portuguese.
\newblock \emph{Corpora}, 12(1):23--54.

\bibitem[{Carlson et~al.(2001)Carlson, Marcu, and Okurovsky}]{carlson-etal-2001-building}
Lynn Carlson, Daniel Marcu, and Mary~Ellen Okurovsky. 2001.
\newblock \href {https://aclanthology.org/W01-1605/} {Building a discourse-tagged corpus in the framework of {R}hetorical {S}tructure {T}heory}.
\newblock In \emph{Proceedings of the Second {SIG}dial Workshop on Discourse and Dialogue}.

\bibitem[{Chatterji et~al.(2025)Chatterji, Cunningham, Deming, Hitzig, Ong, Shan, and Wadman}]{chatterji2025people}
Aaron Chatterji, Thomas Cunningham, David~J Deming, Zoe Hitzig, Christopher Ong, Carl~Yan Shan, and Kevin Wadman. 2025.
\newblock How people use chatgpt.
\newblock Technical report, National Bureau of Economic Research.

\bibitem[{Chistova(2024)}]{chistova-2024-bilingual}
Elena Chistova. 2024.
\newblock \href {https://doi.org/10.18653/v1/2024.findings-acl.577} {Bilingual rhetorical structure parsing with large parallel annotations}.
\newblock In \emph{Findings of the Association for Computational Linguistics: ACL 2024}, pages 9689--9706, Bangkok, Thailand. Association for Computational Linguistics.

\bibitem[{Cole et~al.(2023)Cole, Zhang, Gillick, Eisenschlos, Dhingra, and Eisenstein}]{cole-etal-2023-selectively}
Jeremy Cole, Michael Zhang, Daniel Gillick, Julian Eisenschlos, Bhuwan Dhingra, and Jacob Eisenstein. 2023.
\newblock \href {https://doi.org/10.18653/v1/2023.emnlp-main.35} {Selectively answering ambiguous questions}.
\newblock In \emph{Proceedings of the 2023 Conference on Empirical Methods in Natural Language Processing}, pages 530--543, Singapore. Association for Computational Linguistics.

\bibitem[{Dietz et~al.(2017)Dietz, Verma, Radlinski, and Craswell}]{dietz2017trec}
Laura Dietz, Manisha Verma, Filip Radlinski, and Nick Craswell. 2017.
\newblock Trec complex answer retrieval overview.
\newblock In \emph{TREC}.

\bibitem[{Ethayarajh et~al.(2022)Ethayarajh, Choi, and Swayamdipta}]{pmlr-v162-ethayarajh22a}
Kawin Ethayarajh, Yejin Choi, and Swabha Swayamdipta. 2022.
\newblock Understanding dataset difficulty with $\mathcal{V}$-usable information.
\newblock In \emph{Proceedings of the 39th International Conference on Machine Learning}, volume 162 of \emph{Proceedings of Machine Learning Research}, pages 5988--6008. PMLR.

\bibitem[{Fan et~al.(2019)Fan, Jernite, Perez, Grangier, Weston, and Auli}]{fan2019eli5}
Angela Fan, Yacine Jernite, Ethan Perez, David Grangier, Jason Weston, and Michael Auli. 2019.
\newblock Eli5: Long form question answering.
\newblock In \emph{Proceedings of the 57th annual meeting of the association for computational linguistics}, pages 3558--3567.

\bibitem[{Grice(1975)}]{grice1975logic}
Herbert~P Grice. 1975.
\newblock Logic and conversation.
\newblock In \emph{Speech acts}, pages 41--58. Brill.

\bibitem[{Hu et~al.(2023)Hu, Song, Cho, Wang, Foroosh, and Liu}]{hu-etal-2023-decipherpref}
Yebowen Hu, Kaiqiang Song, Sangwoo Cho, Xiaoyang Wang, Hassan Foroosh, and Fei Liu. 2023.
\newblock \href {https://doi.org/10.18653/v1/2023.emnlp-main.519} {{D}ecipher{P}ref: Analyzing influential factors in human preference judgments via {GPT}-4}.
\newblock In \emph{Proceedings of the 2023 Conference on Empirical Methods in Natural Language Processing}, pages 8344--8357, Singapore. Association for Computational Linguistics.

\bibitem[{Kim et~al.(2023)Kim, Kim, Jeon, Park, and Kang}]{kim-etal-2023-tree}
Gangwoo Kim, Sungdong Kim, Byeongguk Jeon, Joonsuk Park, and Jaewoo Kang. 2023.
\newblock \href {https://doi.org/10.18653/v1/2023.emnlp-main.63} {Tree of clarifications: Answering ambiguous questions with retrieval-augmented large language models}.
\newblock In \emph{Proceedings of the 2023 Conference on Empirical Methods in Natural Language Processing}, pages 996--1009, Singapore. Association for Computational Linguistics.

\bibitem[{Kim et~al.(2021)Kim, Pavlick, Karagol-Ayan, and Ramachandran}]{kim2021linguist}
Najoung Kim, Ellie Pavlick, Burcu Karagol-Ayan, and Deepak Ramachandran. 2021.
\newblock Which linguist invented the lightbulb? presupposition verification for question-answering.
\newblock In \emph{Proceedings of the 59th Annual Meeting of the Association for Computational Linguistics and the 11th International Joint Conference on Natural Language Processing (Volume 1: Long Papers)}, pages 3932--3945.

\bibitem[{Kim et~al.(2024)Kim, Lee, Zhu, Raheja, and Kang}]{kim-etal-2024-threads}
Zae~Myung Kim, Kwang Lee, Preston Zhu, Vipul Raheja, and Dongyeop Kang. 2024.
\newblock \href {https://doi.org/10.18653/v1/2024.acl-long.298} {Threads of subtlety: Detecting machine-generated texts through discourse motifs}.
\newblock In \emph{Proceedings of the 62nd Annual Meeting of the Association for Computational Linguistics (Volume 1: Long Papers)}, pages 5449--5474, Bangkok, Thailand. Association for Computational Linguistics.

\bibitem[{Ko et~al.(2023)Ko, Wu, Dalton, Srinivas, Durrett, and Li}]{ko2023discourse}
Wei-Jen Ko, Yating Wu, Cutter Dalton, Dananjay Srinivas, Greg Durrett, and Junyi~Jessy Li. 2023.
\newblock Discourse analysis via questions and answers: Parsing dependency structures of questions under discussion.
\newblock In \emph{Findings of the Association for Computational Linguistics: ACL 2023}, pages 11181--11195.

\bibitem[{Krishna et~al.(2021)Krishna, Roy, and Iyyer}]{krishna2021hurdles}
Kalpesh Krishna, Aurko Roy, and Mohit Iyyer. 2021.
\newblock Hurdles to progress in long-form question answering.
\newblock In \emph{Proceedings of the 2021 Conference of the North American Chapter of the Association for Computational Linguistics: Human Language Technologies}, pages 4940--4957.

\bibitem[{Kumar et~al.(2025)Kumar, Park, Tsvetkov, Smith, and Hajishirzi}]{kumar-etal-2025-compo}
Sachin Kumar, Chan~Young Park, Yulia Tsvetkov, Noah~A. Smith, and Hannaneh Hajishirzi. 2025.
\newblock \href {https://doi.org/10.18653/v1/2025.naacl-long.419} {{C}om{PO}: Community preferences for language model personalization}.
\newblock In \emph{Proceedings of the 2025 Conference of the Nations of the Americas Chapter of the Association for Computational Linguistics: Human Language Technologies (Volume 1: Long Papers)}, pages 8246--8279, Albuquerque, New Mexico. Association for Computational Linguistics.

\bibitem[{Kwiatkowski et~al.(2019)Kwiatkowski, Palomaki, Redfield, Collins, Parikh, Alberti, Epstein, Polosukhin, Devlin, Lee, Toutanova, Jones, Kelcey, Chang, Dai, Uszkoreit, Le, and Petrov}]{kwiatkowski-etal-2019-natural}
Tom Kwiatkowski, Jennimaria Palomaki, Olivia Redfield, Michael Collins, Ankur Parikh, Chris Alberti, Danielle Epstein, Illia Polosukhin, Jacob Devlin, Kenton Lee, Kristina Toutanova, Llion Jones, Matthew Kelcey, Ming-Wei Chang, Andrew~M. Dai, Jakob Uszkoreit, Quoc Le, and Slav Petrov. 2019.
\newblock \href {https://doi.org/10.1162/tacl_a_00276} {Natural questions: A benchmark for question answering research}.
\newblock \emph{Transactions of the Association for Computational Linguistics}, 7:452--466.

\bibitem[{Maekawa et~al.(2024)Maekawa, Hirao, Kamigaito, and Okumura}]{maekawa-etal-2024-obtain}
Aru Maekawa, Tsutomu Hirao, Hidetaka Kamigaito, and Manabu Okumura. 2024.
\newblock \href {https://doi.org/10.18653/v1/2024.eacl-long.171} {Can we obtain significant success in {RST} discourse parsing by using large language models?}
\newblock In \emph{Proceedings of the 18th Conference of the European Chapter of the Association for Computational Linguistics (Volume 1: Long Papers)}, pages 2803--2815, St. Julian{'}s, Malta. Association for Computational Linguistics.

\bibitem[{Majumder et~al.(2021)Majumder, Rao, Galley, and McAuley}]{majumder-etal-2021-ask}
Bodhisattwa~Prasad Majumder, Sudha Rao, Michel Galley, and Julian McAuley. 2021.
\newblock \href {https://doi.org/10.18653/v1/2021.naacl-main.340} {Ask what{'}s missing and what{'}s useful: Improving clarification question generation using global knowledge}.
\newblock In \emph{Proceedings of the 2021 Conference of the North American Chapter of the Association for Computational Linguistics: Human Language Technologies}, pages 4300--4312, Online. Association for Computational Linguistics.

\bibitem[{Malaviya et~al.(2024)Malaviya, Lee, Chen, Sieber, Yatskar, and Roth}]{malaviya-etal-2024-expertqa}
Chaitanya Malaviya, Subin Lee, Sihao Chen, Elizabeth Sieber, Mark Yatskar, and Dan Roth. 2024.
\newblock \href {https://doi.org/10.18653/v1/2024.naacl-long.167} {{E}xpert{QA}: Expert-curated questions and attributed answers}.
\newblock In \emph{Proceedings of the 2024 Conference of the North American Chapter of the Association for Computational Linguistics: Human Language Technologies (Volume 1: Long Papers)}, pages 3025--3045, Mexico City, Mexico. Association for Computational Linguistics.

\bibitem[{Mann and Thompson(1987)}]{mann1987rhetorical}
William~C Mann and Sandra~A Thompson. 1987.
\newblock Rhetorical structure theory: A theory of text organization.
\newblock Technical report, University of Southern California, Information Sciences Institute Los Angeles.

\bibitem[{Meng et~al.(2023)Meng, Pan, Cao, and Kan}]{meng-etal-2023-followupqg}
Yan Meng, Liangming Pan, Yixin Cao, and Min-Yen Kan. 2023.
\newblock \href {https://doi.org/10.18653/v1/2023.ijcnlp-main.17} {{F}ollowup{QG}: Towards information-seeking follow-up question generation}.
\newblock In \emph{Proceedings of the 13th International Joint Conference on Natural Language Processing and the 3rd Conference of the Asia-Pacific Chapter of the Association for Computational Linguistics (Volume 1: Long Papers)}, pages 252--271, Nusa Dua, Bali. Association for Computational Linguistics.

\bibitem[{Min et~al.(2020)Min, Michael, Hajishirzi, and Zettlemoyer}]{min-etal-2020-ambigqa}
Sewon Min, Julian Michael, Hannaneh Hajishirzi, and Luke Zettlemoyer. 2020.
\newblock \href {https://doi.org/10.18653/v1/2020.emnlp-main.466} {{A}mbig{QA}: Answering ambiguous open-domain questions}.
\newblock In \emph{Proceedings of the 2020 Conference on Empirical Methods in Natural Language Processing (EMNLP)}, pages 5783--5797, Online. Association for Computational Linguistics.

\bibitem[{Mu{\~n}oz-Ortiz et~al.(2024)Mu{\~n}oz-Ortiz, G{\'o}mez-Rodr{\'\i}guez, and Vilares}]{munoz2024contrasting}
Alberto Mu{\~n}oz-Ortiz, Carlos G{\'o}mez-Rodr{\'\i}guez, and David Vilares. 2024.
\newblock Contrasting linguistic patterns in human and llm-generated news text.
\newblock \emph{Artificial Intelligence Review}, 57(10):265.

\bibitem[{Nakano et~al.(2021)Nakano, Hilton, Balaji, Wu, Ouyang, Kim, Hesse, Jain, Kosaraju, Saunders et~al.}]{nakano2021webgpt}
Reiichiro Nakano, Jacob Hilton, Suchir Balaji, Jeff Wu, Long Ouyang, Christina Kim, Christopher Hesse, Shantanu Jain, Vineet Kosaraju, William Saunders, et~al. 2021.
\newblock Webgpt: Browser-assisted question-answering with human feedback.
\newblock \emph{arXiv preprint arXiv:2112.09332}.

\bibitem[{Namuduri et~al.(2025)Namuduri, Wu, Zheng, Wadhwa, Durrett, and Li}]{namuduriqudsim}
Ramya Namuduri, Yating Wu, Anshun~Asher Zheng, Manya Wadhwa, Greg Durrett, and Junyi~Jessy Li. 2025.
\newblock Qudsim: Quantifying discourse similarities in llm-generated text.
\newblock In \emph{Second Conference on Language Modeling}.

\bibitem[{Ouyang et~al.(2022)Ouyang, Wu, Jiang, Almeida, Wainwright, Mishkin, Zhang, Agarwal, Slama, Ray et~al.}]{ouyang2022training}
Long Ouyang, Jeffrey Wu, Xu~Jiang, Diogo Almeida, Carroll Wainwright, Pamela Mishkin, Chong Zhang, Sandhini Agarwal, Katarina Slama, Alex Ray, et~al. 2022.
\newblock Training language models to follow instructions with human feedback.
\newblock \emph{Advances in neural information processing systems}, 35:27730--27744.

\bibitem[{Poddar et~al.(2025)Poddar, Koley, Misra, Ganguly, and Ghosh}]{poddar2025brevity}
Soham Poddar, Paramita Koley, Janardan Misra, Niloy Ganguly, and Saptarshi Ghosh. 2025.
\newblock Brevity is the soul of sustainability: Characterizing llm response lengths.
\newblock In \emph{Findings of the Association for Computational Linguistics: ACL 2025}, pages 21848--21864.

\bibitem[{Ramshaw and Marcus(1995)}]{ramshaw-marcus-1995-text}
Lance Ramshaw and Mitch Marcus. 1995.
\newblock \href {https://aclanthology.org/W95-0107/} {Text chunking using transformation-based learning}.
\newblock In \emph{Third Workshop on Very Large Corpora}.

\bibitem[{Reinhart et~al.(2025)Reinhart, Markey, Laudenbach, Pantusen, Yurko, Weinberg, and Brown}]{reinhart2025llms}
Alex Reinhart, Ben Markey, Michael Laudenbach, Kachatad Pantusen, Ronald Yurko, Gordon Weinberg, and David~West Brown. 2025.
\newblock Do llms write like humans? variation in grammatical and rhetorical styles.
\newblock \emph{Proceedings of the National Academy of Sciences}, 122(8):e2422455122.

\bibitem[{Roberts(2012)}]{roberts2012information}
Craige Roberts. 2012.
\newblock Information structure: Towards an integrated formal theory of pragmatics.
\newblock \emph{Semantics and pragmatics}, 5:6--1.

\bibitem[{Rogers et~al.(2023)Rogers, Gardner, and Augenstein}]{rogers2023qa}
Anna Rogers, Matt Gardner, and Isabelle Augenstein. 2023.
\newblock Qa dataset explosion: A taxonomy of nlp resources for question answering and reading comprehension.
\newblock \emph{ACM Computing Surveys}, 55(10):1--45.

\bibitem[{Searle(2014)}]{searle2014speech}
John Searle. 2014.
\newblock What is a speech act?
\newblock In \emph{Philosophy in America}, pages 221--239. Routledge.

\bibitem[{Searle(1969)}]{searle1969speech}
John~R Searle. 1969.
\newblock \emph{Speech acts: An essay in the philosophy of language}.
\newblock Cambridge university press.

\bibitem[{Shelby et~al.(2025)Shelby, Diaz, and Prabhakaran}]{shelby2025taxonomy}
Renee Shelby, Fernando Diaz, and Vinodkumar Prabhakaran. 2025.
\newblock Taxonomy of user needs and actions.
\newblock \emph{arXiv preprint arXiv:2510.06124}.

\bibitem[{Sperber and Wilson(1986)}]{sperber1986relevance}
Dan Sperber and Deirdre Wilson. 1986.
\newblock \emph{Relevance: Communication and cognition}, volume 142.
\newblock Harvard University Press Cambridge, MA.

\bibitem[{Srikanth et~al.(2025)Srikanth, Rudinger, and Boyd-Graber}]{srikanth2025no}
Neha Srikanth, Rachel Rudinger, and Jordan~Lee Boyd-Graber. 2025.
\newblock No questions are stupid, but some are poorly posed: Understanding poorly-posed information-seeking questions.
\newblock In \emph{Proceedings of the 63rd Annual Meeting of the Association for Computational Linguistics (Volume 1: Long Papers)}, pages 3182--3199.

\bibitem[{Srikanth et~al.(2024)Srikanth, Sarkar, Mane, Aparicio, Nguyen, Rudinger, and Boyd-Graber}]{srikanth-etal-2024-pregnant}
Neha Srikanth, Rupak Sarkar, Heran Mane, Elizabeth Aparicio, Quynh Nguyen, Rachel Rudinger, and Jordan Boyd-Graber. 2024.
\newblock \href {https://doi.org/10.18653/v1/2024.naacl-long.403} {Pregnant questions: The importance of pragmatic awareness in maternal health question answering}.
\newblock In \emph{Proceedings of the 2024 Conference of the North American Chapter of the Association for Computational Linguistics: Human Language Technologies (Volume 1: Long Papers)}, pages 7253--7268, Mexico City, Mexico. Association for Computational Linguistics.

\bibitem[{Stelmakh et~al.(2022)Stelmakh, Luan, Dhingra, and Chang}]{stelmakh-etal-2022-asqa}
Ivan Stelmakh, Yi~Luan, Bhuwan Dhingra, and Ming-Wei Chang. 2022.
\newblock \href {https://doi.org/10.18653/v1/2022.emnlp-main.566} {{ASQA}: Factoid questions meet long-form answers}.
\newblock In \emph{Proceedings of the 2022 Conference on Empirical Methods in Natural Language Processing}, pages 8273--8288, Abu Dhabi, United Arab Emirates. Association for Computational Linguistics.

\bibitem[{Stengel-Eskin et~al.(2021)Stengel-Eskin, Guallar-Blasco, and Van~Durme}]{stengel2021human}
Elias Stengel-Eskin, Jimena Guallar-Blasco, and Benjamin Van~Durme. 2021.
\newblock Human-model divergence in the handling of vagueness.
\newblock In \emph{Proceedings of the 1st Workshop on Understanding Implicit and Underspecified Language}, pages 43--57.

\bibitem[{Stolcke et~al.(2000)Stolcke, Ries, Coccaro, Shriberg, Bates, Jurafsky, Taylor, Martin, Van Ess-Dykema, and Meteer}]{stolcke-etal-2000-dialogue}
Andreas Stolcke, Klaus Ries, Noah Coccaro, Elizabeth Shriberg, Rebecca Bates, Daniel Jurafsky, Paul Taylor, Rachel Martin, Carol Van Ess-Dykema, and Marie Meteer. 2000.
\newblock \href {https://aclanthology.org/J00-3003/} {Dialogue act modeling for automatic tagging and recognition of conversational speech}.
\newblock \emph{Computational Linguistics}, 26(3):339--374.

\bibitem[{Taylor(1962)}]{taylor1962process}
Robert~S Taylor. 1962.
\newblock The process of asking questions.
\newblock \emph{American documentation}, 13(4):391--396.

\bibitem[{Teufel and Moens(2002)}]{teufel2002summarizing}
Simone Teufel and Marc Moens. 2002.
\newblock Summarizing scientific articles: experiments with relevance and rhetorical status.
\newblock \emph{Computational linguistics}, 28(4):409--445.

\bibitem[{Tsvilodub et~al.(2023)Tsvilodub, Franke, Hawkins, and Goodman}]{tsvilodub2023overinformative}
Polina Tsvilodub, Michael Franke, Robert~D Hawkins, and Noah~D Goodman. 2023.
\newblock Overinformative question answering by humans and machines.
\newblock \emph{arXiv preprint arXiv:2305.07151}.

\bibitem[{Wang et~al.(2025)Wang, Zhou, Zheng, Huang, Luo, Luo, and Bai}]{wang-etal-2025-uncovering}
Yini Wang, Xian Zhou, Shengan Zheng, Linpeng Huang, Zhunchen Luo, Wei Luo, and Xiaoying Bai. 2025.
\newblock \href {https://doi.org/10.18653/v1/2025.emnlp-main.416} {Uncovering argumentative flow: A question-focus discourse structuring framework}.
\newblock In \emph{Proceedings of the 2025 Conference on Empirical Methods in Natural Language Processing}, pages 8241--8259, Suzhou, China. Association for Computational Linguistics.

\bibitem[{Wu et~al.(2023)Wu, Mangla, Durrett, and Li}]{wu2023qudeval}
Yating Wu, Ritika Mangla, Greg Durrett, and Junyi~Jessy Li. 2023.
\newblock Qudeval: The evaluation of questions under discussion discourse parsing.
\newblock In \emph{Proceedings of the 2023 Conference on Empirical Methods in Natural Language Processing}, pages 5344--5363.

\bibitem[{Xu et~al.(2022)Xu, Li, and Choi}]{xu2022we}
Fangyuan Xu, Junyi~Jessy Li, and Eunsol Choi. 2022.
\newblock How do we answer complex questions: Discourse structure of long-form answers.
\newblock In \emph{Proceedings of the 60th Annual Meeting of the Association for Computational Linguistics (Volume 1: Long Papers)}, pages 3556--3572.

\bibitem[{Xu et~al.(2023)Xu, Song, Iyyer, and Choi}]{xu2023critical}
Fangyuan Xu, Yixiao Song, Mohit Iyyer, and Eunsol Choi. 2023.
\newblock A critical evaluation of evaluations for long-form question answering.
\newblock In \emph{Proceedings of the 61st Annual Meeting of the Association for Computational Linguistics (Volume 1: Long Papers)}, pages 3225--3245.

\bibitem[{Yu et~al.(2023)Yu, Min, Zettlemoyer, and Hajishirzi}]{yu-etal-2023-crepe}
Xinyan Yu, Sewon Min, Luke Zettlemoyer, and Hannaneh Hajishirzi. 2023.
\newblock \href {https://doi.org/10.18653/v1/2023.acl-long.583} {{CREPE}: Open-domain question answering with false presuppositions}.
\newblock In \emph{Proceedings of the 61st Annual Meeting of the Association for Computational Linguistics (Volume 1: Long Papers)}, pages 10457--10480, Toronto, Canada. Association for Computational Linguistics.

\bibitem[{Zhang et~al.(2017)Zhang, Culbertson, and Paritosh}]{zhang2017characterizing}
Amy Zhang, Bryan Culbertson, and Praveen Paritosh. 2017.
\newblock Characterizing online discussion using coarse discourse sequences.
\newblock In \emph{proceedings of the international AAAI conference on web and social media}, volume~11, pages 357--366.

\bibitem[{Zhang et~al.(2025)Zhang, Knox, and Choi}]{ICLR2025_97e2df4b}
Michael Zhang, W.~Bradley Knox, and Eunsol Choi. 2025.
\newblock \href {https://proceedings.iclr.cc/paper_files/paper/2025/file/97e2df4bb8b2f1913657344a693166a2-Paper-Conference.pdf} {Modeling future conversation turns to teach llms to ask clarifying questions}.
\newblock In \emph{International Conference on Learning Representations}, volume 2025, pages 60722--60742.

\end{thebibliography}

\appendix

\clearpage
\begin{table}[t!]
\centering
\small
\begin{tabular}{ll}
\toprule
\textbf{Relations} \\
\midrule
Elaboration, Attribution, Joint, Same-Unit, Explanation, \\
Enablement, Background, Evaluation, Cause, Contrast, \\
Temporal, Comparison, Topic-Change, Manner-Means, \\
Textual-Organization, Condition, Summary, Topic-Comment \\
\bottomrule
\end{tabular}
\caption{RST-DT relations (\small\url{https://huggingface.co/tchewik/isanlp_rst_v3/blob/rstdt/relation_table.txt}.)}
\label{tab:rst-relations}
\end{table}

\begin{table}[t!]
\centering
\small
\begin{tabular}{cc}
\toprule
\textbf{Relation} & \textbf{Nuclearity} \\
\midrule
Contrast & NN \\
Comparison & NN \\
Topic-Change & NN, NS, SN \\
Evaluation & NS, SN, NN \\
Summary & NN, NS, SN \\
Background & NS, SN \\
\bottomrule
\end{tabular}
\caption{Boundary relations used for RST-based discourse segmentation.}
\label{tab:boundary-relations}
\end{table}

\begin{table*}[t]
\centering
\small
\resizebox{2.08\columnwidth}{!}{
\begin{tabular}{p{3cm}l}
\toprule
\textbf{Subreddit} & \textbf{Description} \\
\midrule
\subreddit{AskHistorians} & Users ask questions about history and receive detailed answers from knowledgeable historians. \\
\subreddit{NoStupidQuestions} & Users casually ask questions about general knowledge or any topic and receive answers. \\
\subreddit{AskEconomics} & Users ask questions about economic theory, research, and policy and receive answers grounded in economic theory and empirical research. \\
\subreddit{asklinguistics} & Users ask questions about linguistics and receive answers from knowledgeable linguists. \\
\subreddit{history} & Users discuss historical topics. \\
\subreddit{OutOfTheLoop} & Users ask questions about current events, pop culture, or internet trends they feel out of the loop on and receive answers. \\
\subreddit{ScienceBasedParenting} & Users ask questions about parenting and discuss research and science-based guidance. \\
\subreddit{beyondthebump} & Users discuss pregnancy, childbirth, and early parenting experiences. \\
\subreddit{explainlikeimfive} & Users ask questions about any topic and receive answers that simplify complex concepts in a way that is accessible for laypeople.\\
\bottomrule
\end{tabular}}
\caption{Descriptions of the subreddits included in our dataset.}
\label{tab:subreddits}
\end{table*}

\section{Data Filtering}
\label{appendix:question-selection}
We use the \texttt{PushShift}\footnote{\url{https://github.com/Watchful1/PushshiftDumps}} dump of our nine subreddits, which contains posts till the end of 2022.
Our study is concerned with information-seeking questions and their answers, so we apply a series of filters accordingly to identify those posts.
\paragraph{Title-Based Filtering.} We filter out posts with no title, less than five upvotes, and less than four tokens. 
To identify \textit{information-seeking} questions instead of community-seeking or other commentary, we use surface form features of post titles, filtering out posts whose titles do not contain a \textit{wh}-question word or that do not end with a question mark.
We also remove posts with Reddit-specific terms (\textit{subreddit}, \textit{redditor}, \textit{upvote} or \textit{karma}) and posts that have multiple questions or statements in the title.
We run a profanity and NSFW detection model (\url{https://github.com/dimitrismistriotis/alt-profanity-check}) on post titles, removing posts above probability $0.8$.
We also filter out posts that contain first-person pronouns, relationship terms (e.g. \textit{my husband}), deictic references combined with first-person pronouns, or validation seeking phrases (``is this normal'', ``does anyone else'', ``is this okay'').
These filters help narrow our pool to \textit{information-seeking} questions.

\paragraph{Thread and Comment-Based Filtering.} We want to ensure that our dataset has high quality comments, since we are studying answerer rhetorical strategy. 
We only consider top-level comments of a thread, and remove top-level comments with less than three upvotes.
We filter out posts with more than \texttt{max} comments to avoid threads with unusually high engagement, which tend to be controversial, opinion-based, or discussion-oriented rather than information-seeking.
We also filter out posts with less than \texttt{min} comments since we want enough answers for the same question to compare different strategies that answerers may use. 
We set \texttt{min\_comments} to 5 for most subreddits and 4 for \subreddit{ScienceBasedParenting} and \texttt{max\_comments} to 12 for \subreddit{AskHistorians}, \subreddit{history}, \subreddit{OutOfTheLoop}, and \subreddit{beyondthebump}, 15 for \subreddit{asklinguistics} and \subreddit{explainlikeimfive}, 18 for \subreddit{NoStupidQuestions}, 20 for \subreddit{ScienceBasedParenting}, and 30 for \subreddit{AskEconomics}.

\paragraph{Sampling.} After filtering posts from each subreddit according to all the filters above, we sample 300 questions and their top-level comments from each, yielding 2,700 questions and 18,968 answers.

\section{Preprocessing Answers with Rhetorical Structure Theory}
\label{appendix:rst}
Rhetorical structure theory~\cite{mann1987rhetorical}, or RST, explains the coherence of a text by expressing how components of a text are related to each other.
RST represents a text $t$ as a binary tree. The root represents the whole text.
Leaves are \textit{elementary discourse units} (EDUs), which are clauses in $t$.
Leaves are recursively connected through intermediate nodes which express the relation \textit{between} children (Table~\ref{tab:rst-relations}) as well as the distinction between the
primary (``nucleus'') and supporting (``satellite'') child
(Fig.~\ref{fig:rst-parsing}).

\paragraph{Coarsening EDUs to produce action segments.} Elementary discourse units in RST are often sentence \textit{clauses} (see Figure~\ref{fig:rst-parsing}) and hence are too short to serve as segments that we can label with a higher-level discourse relation to the question.
We opt to merge EDUs that we suspect are part of the same discourse act to form coarser segments that we label with an act.
To do this, we define \textit{boundary relations} (Table~\ref{tab:boundary-relations}): a subset of the full set of RST relations (Table~\ref{tab:rst-relations}) that are likely to signal a rhetorical shift between a node's children. Boundary relations are defined as relation--nuclearity pairs, since the same relation (e.g., \textit{Evaluation}) may signal a shift under some nuclearity configurations but not others.

Then, we execute the procedure detailed in Algorithm~\ref{alg:segment} using these boundary relations.
Starting at the root, we recursively traverse the RST tree to decide where to split. 
At an intermediate node $v$, we first recurse into both children to obtain a list of text spans $L$ and $R$. 
If the relation--nuclearity pair of $v$ is a boundary relation, we split $v$'s children into separate segments, except for the \textit{Background} relation, where we require both sides to be sufficiently large ($\geq k$ EDUs) to avoid creating trivially small segments from brief contextualizing clauses.
If the size check fails, the entire subtree is collapsed into a single span. 
Even if a relation is not a boundary relation, we still check whether either child was split deeper in the tree.
If so, we preserve those deeper splits, but otherwise, we collapse the subtree into one span. 
Our procedure produces a list of text spans, referred to as \textbf{action segments}, which we suspect correspond to distinct discourse acts that we label in \S\ref{subsec:actions}.

\begin{algorithm}[t]
\caption{RST-Based Discourse Segmentation}
\label{alg:segment}
\SetAlgoLined
\KwIn{RST tree node $v$, boundary relations $\mathcal{B}$, minimum span size $k=3$}
\KwOut{List of EDU spans (each span is a list of EDUs)}
\SetKwFunction{FGetSpans}{GetSpans}
\SetKwFunction{FGetLeaves}{GetLeaves}
\SetKwProg{Fn}{Function}{:}{}
\Fn{\FGetSpans{$v$, $\mathcal{B}$, $k$}}{
    \lIf{$v$ is a leaf}{\Return $[[v]]$}
    $L \leftarrow$ \FGetSpans{$v.\textit{left}$, $\mathcal{B}$, $k$}\;
    $R \leftarrow$ \FGetSpans{$v.\textit{right}$, $\mathcal{B}$, $k$}\;
    \If{$v.\textit{relation}$\texttt{\_}$v.\textit{nuclearity} \in \mathcal{B}$}{
        \tcp{$\|L\|, \|R\|$ = total number of EDUs in $L$, $R$}
        \lIf{$v.\textit{relation} \neq \textit{Background}$ \textbf{or}
            $(\|L\| \geq k$ \textbf{and} $\|R\| \geq k)$}{\Return $L + R$}
        \lElse{\Return $[\,\text{\FGetLeaves}(v)\,]$}
    }
    \tcp{Not a boundary relation; preserve any splits from deeper in the tree}
    \lIf{$|L| + |R| > 2$}{\Return $L + R$}
    \Return $[\,\text{\FGetLeaves}(v)\,]$\;
}
\BlankLine
\tcp{\text{\FGetLeaves}$(v)$ returns all leaf EDUs of subtree $v$ as a single flat list}
\end{algorithm}

\section{Discourse Act Ontology Construction}
\label{appendix:ontology-construction}
We developed our ontology iteratively, starting from the six sentence-level
roles of \citet{xu2022we}. 
These roles were designed to capture how answers
\emph{organize} information. However, our goal is to characterize the \textit{function} each segment serves with respect to an answerer's rhetorical \textit{strategy}.
Labeling RST-derived segments from the Reddit community \subreddit{r/NoStupidQuestions} with these roles, we found that, from a strategy perspective, roles like \textit{summary} and \textit{example} are realizations of a broader ``answering'' move, while their \textit{organization} role spans both framing and substantive acts.
Additionally, their role set did not include representation for the interrogative acts we observe in \texttt{r/NoStupidQuestions}.
We therefore restructured these roles into functional families (answer question (AQ), seek information (SI), comment on the question (CQ), redirect the question (RQ), and no-op (NO)), and then repeatedly sampled new QA pairs, adding acts to each of these families when no existing act was appropriate or merging acts together until ten consecutive pairs were sampled without any modifications.
We then stress-tested on other communities (e.g. \subreddit{AskHistorians}) which 
yielded further acts (e.g. Cite External Source) and surfaced overloaded
acts that required splits (e.g. Counter-Question from Redirect
Formulation).
Again, we randomly and uniformly sampled QA pairs from our nine different Reddit communities (\S\ref{sec:communities}) until ten consecutive samples required no ontology modifications. 
We treated all pairs used for the development of our ontology as ``training samples'' and discard them from any other experiments and analysis we do throughout the work.

\section{Discourse Act Labeling}
\label{appendix:act-labeling}

We label the segments produced from Algorithm~\ref{alg:segment} using \model{gpt-4.1-2025-0414} using Prompt~\ref{prompt:discourse-act-tagging}.

\
\begin{prompt}[title={Prompt \thetcbcounter: Discourse Act Tagging}, label=prompt:discourse-act-tagging]
{\ttfamily
\colorbox{cyan!20}{System:} You are an expert discourse analyst trying to understand how people answer questions on Reddit. You will analyze answers by tagging each text segment from a Reddit answer with a discourse action.

\textbf{Rules}\\
1. \textbf{Select EXACTLY ONE action\_id per segment or subsegment.}\\
2. If the current segment continues the previous action, reuse the previous action\_id.\\
3. If a new rhetorical move begins, select the appropriate new action\_id.\\
4. If no action fits, use ``NONE''.

\textbf{Subsegment Labeling}\\
Each segment you receive was produced by a discourse parser. You will also be shown the subsegments (sentences) that make up the segment. If all subsegments serve the same discourse function, return a single-element array with one action\_id. If different subsegments serve \textbf{different discourse functions}, return an array with one entry per subsegment, each with its subsegment\_index and action\_id.

Common patterns worth splitting:\\
- Background/reasoning subsegments followed by an answer subsegment\\
- An answer subsegment followed by a redirect or recommendation\\
- A presupposition rejection followed by an alternative answer\\

Do NOT split when:\\
- A subsegment contains light framing for the next (e.g., ``So basically,'' followed by an answer $\rightarrow$ single Assert Answer)\\
- The difference is just emphasis vs.\ substance within the same move\\

\textbf{Caveats and Task Nuances}\\
1. Consider the expected answer type of the question when labeling actions. Responding to ``Where can I find X'' with a website recommendation is an ``Answer the Question'' action, not a ``Direct to Resource'' action. If the resource is the answer itself, label it as ``Assert Answer''. If the resource is suggested as additional reading, use Direct to Resource.\\
2. When a segment contains both an answer and supporting reasoning: if subsegments are provided and the answer and reasoning fall in \textbf{different subsegments}, split them. If they are in the \textbf{same subsegment} (tightly integrated), label it as ``Provide Reasoning or Justification'' if the justification is non-trivial, otherwise ``Assert Answer''.\\
3. When a segment explains WHY something is the case, determine what it is explaining:\\
- If it explains why an \textit{answer} is correct $\rightarrow$ ``Provide Reasoning or Justification''\\
- If it explains why a \textit{premise of the question} is wrong $\rightarrow$ ``Reject Presupposition''\\
Example: For ``Why is the sky blue?'', the segment ``Because of Rayleigh scattering'' is justification. For ``What's the best liver detox cleanse?'', the segment ``The concept of `detoxing' your liver is misleading---your liver already filters toxins continuously'' is rejecting the presupposition.\\
4. Sharing a personal anecdote or experience is ``Provide Example'', NOT ``Provide Background.'' Background sets up context, frameworks, or history \textit{before} answering. Examples use concrete cases (including personal ones) to \textit{support or illustrate} an answer.\\
- ``I have a doctorate, and sometimes introduce myself as Dr.'' $\rightarrow$ Provide Example\\
- ``The use of honorifics has a long and contested history in academia.'' $\rightarrow$ Provide Background\\
5. When a segment follows a recommendation and provides supporting information, ask: does it explain why the recommendation is good \textit{in terms of the original question}, or does it answer a \textit{different} question?\\
- If it explains why the recommendation addresses the original question $\rightarrow$ ``Provide Reasoning or Justification''\\
- If it introduces new information that answers a tangentially related but different question $\rightarrow$ ``Answer a Question or Interpretation outside of Interpretation Space''\\
6. When a segment invokes an external source (study, statistic, law, quote, expert consensus) to support a claim, use ``Cite External Source''---NOT ``Provide Example'' or ``Provide Reasoning.'' The key test: does the credibility derive from an independently verifiable external source, or from the answerer's own experience/logic?\\
- ``A 2019 Lancet study found no significant effect.'' $\rightarrow$ Cite External Source\\
- ``I saw the same thing happen at my last job.'' $\rightarrow$ Provide Example\\
- ``That's because the compiler needs type info at compile time.'' $\rightarrow$ Provide Reasoning\\

\textbf{Action Ontology}\\
\{ontology\}

\textbf{Output Format}\\
Always respond with ONLY a JSON array. No explanation, no reasoning, no commentary.

Single action for whole segment:\\
{[}\{"action\_id": "action\_AQ\_assert\_answer"\}{]}\\

Distinct actions per subsegment:\\
{[}\{"subsegment\_index": 0, "action\_id": "action\_CQ\_reject\_presupposition"\}, \{"subsegment\_index": 1, "action\_id": "action\_AQ\_assert\_answer"\}{]}\\

When no action fits:\\
{[}\{"action\_id": "NONE"\}{]}

\colorbox{yellow!20}{User:}

\textbf{Question}\\
\{question\}

\textbf{Full Answer}\\
\{answer\}

\textbf{Previous Segment} action=``\{prev\_label\}''\\
\{segment\_prev\}

\textbf{Current Segment}\\
\{segment\}

\textbf{Subsegments}\\
\{subsegments\}

Respond with ONLY a JSON array.
}
\end{prompt}

We observe fairly low usage rates of the label \texttt{NONE} by our automatic act labeler. These rates, stratified across our different communities, are shown in Table~\ref{tab:none_rates}.

\begin{table}[t]
\centering
\small
\resizebox{\columnwidth}{!}{
\begin{tabular}{lrrr}
\toprule
\textbf{Subreddit} & \textbf{Segments} & \textbf{\textsc{None}} & \textbf{\textsc{None} Rate} \\
\midrule
\subreddit{AskHistorians}         & 10,619 & 332 & 3.13\% \\
\subreddit{explainlikeimfive}     &  9,190 & 224 & 2.44\% \\
\subreddit{beyondthebump}         &  6,950 &  75 & 1.08\% \\
\subreddit{NoStupidQuestions}     &  5,108 &  51 & 1.00\% \\
\subreddit{asklinguistics}        &  6,701 & 148 & 2.21\% \\
\subreddit{ScienceBasedParenting} & 11,255 & 189 & 1.68\% \\
\subreddit{history}               &  7,442 & 163 & 2.19\% \\
\subreddit{OutOfTheLoop}          &  8,205 & 190 & 2.32\% \\
\subreddit{AskEconomics}          &  8,511 & 308 & 3.62\% \\
\midrule
Total                             & 73,981 & 1,680 & 2.27\% \\
\bottomrule
\end{tabular}}
\caption{Discourse act \textsc{None} rates by community. Low rates across
all nine communities indicate that our ontology covers the discourse acts
present in answers. We inspect \textsc{None} segments, and attribute them
to RST parsing errors rather than ontology gaps (\S\ref{appendix:ontology-construction}).}
\label{tab:none_rates}
\end{table}

\section{Annotation Guidelines}
The annotation guidelines shown to annotators are below. Annotators are not shown any PII: they are not shown the post ID or the user ID of the asker or the commentor. The only information they are shown is the name of the subreddit in which the question is asked.
Both annotators have graduate degrees in Computer Science or linguistics and are native English speakers.
Annotators were recruited from Upwork and through the department.

\label{appendix:guidelines}
\subsection{Overview}
There are many ways to respond to a question when asked. 
An answerer may choose to ask a follow-up question, probe for some more context, or simply answer by making their best guess about the information needs of the user. Our goal is to identify the communicative moves an answerer makes and, where applicable, which interpretation of the question each move addresses.
Each annotation instance consists of a question Q, an answer A that has been pre-segmented into text spans, and (optionally) a set of candidate interpretations of Q. 
Your task is to assign one discourse act label from the ontology (Fig.~\ref{fig:action-ontology}) to each segment, and for eligible acts, to indicate which interpretation the segment addresses.

\subsection{Task Description}
For each segment in the answer:
\begin{enumerate}
    \item Read the segment in the context of the full answer and the original question Q.
    \item Assign exactly one discourse act label from the ontology. If a segment appears to serve multiple functions, choose the primary or dominant one. 
    \item If the assigned act is marked as interpretation-eligible (starred), select the interpretation from the provided interpretation space that the segment most closely addresses. If no listed interpretation fits, mark it as ``other.''
\end{enumerate}

\subsection{Discourse Act Ontology}
We provided annotators with the same content as in Figure~\ref{fig:action-ontology} as well as a handful of examples.

\subsection{Interpretation Labeling}
Questions often admit multiple valid readings. 
For example, ``Can language be described by math?'' could be interpreted as asking about formal language theory, computational linguistics, or the philosophy of mathematical description.
When a discourse act is interpretation-eligible (starred), you must select which interpretation from the provided set the segment addresses. 
If a segment addresses multiple interpretations, choose the primary one. If it does not clearly map to any listed interpretation, mark \texttt{NONE}.
Acts that are NOT interpretation-eligible (e.g., Reject Presupposition, Presentational, Non-Answer) engage with the question’s framing or serve a structural role rather than adopting a particular reading. 
Do not assign an interpretation to these.

\subsection{Decision Rules and Caveats}
\begin{itemize}
    \item Reject Presupposition vs. Provide Reasoning: If the segment explains why a premise of the question is wrong, use Reject Presupposition. If it explains why an answer is correct, use Provide Reasoning or Justification.
    \item Provide Background vs. Assert Answer: Background situates the question within a broader context without directly answering. If the segment makes a direct claim that resolves the question, prefer Assert Answer.
    \item Cite External Source vs. Provide Example: If credibility derives from the authority of the cited source, use Cite External Source. If a concrete case is offered to illustrate a point (including personal experience), use Provide Example.
    \item Cite External Source vs. Direct to Resource: If the answerer uses content from a source to support their claim in the answer, use Cite External Source. If they redirect the asker to go read or consult the resource themselves, use Direct to Resource.
    \item Comment on Question vs. Presentational: If the remark engages substantively with the question’s nature or difficulty, use Comment on Question. If it is purely filler (“Great question!”), use Presentational.
    \item Redirect Formulation vs. Answer outside Interpretation Space: If the segment explicitly proposes a better way to ask the question, use Redirect Formulation. If it drifts to answering a related but different question without flagging the reframe, use Answer outside Interpretation Space.
    \item Counter-Question vs. Clarification: Counter-Questions challenge the premise or assumptions of the question. Clarifications simply request more information to better answer it.
\end{itemize}

\section{Interpretation Labeling}
\begin{prompt}[title={Prompt \thetcbcounter: Interpretation Generation}, label=prompt:interpretation-generation]
{\ttfamily
\colorbox{cyan!20}{System:} Users in a question answering community typically try to express a need for information through a question. Sometimes, from the language of their question alone, it is not clear what their exact information need is. This leads to many distinct interpretations of their question, each representing different information needs. You will be given a \textbf{question} asked in a specific online community that may have many distinct interpretations. Your task is to output those interpretations as unambiguous distinct questions.

\{subreddit\_context\}

\textbf{Critical Rules}\\
- Each interpretation must be a \textbf{different plausible reading} of the SAME question --- a different thing the user could have MEANT by their words.\\
- Do NOT generate sub-questions, follow-up questions, related questions, or questions that explore different aspects of the topic.\\
- Ask yourself: ``Could the user have typed this exact question while meaning THIS?'' If the answer is no, it is not a valid interpretation.\\
- Interpretations should differ in WHAT the user is asking, not provide additional angles on the same clear question.\\

If the user's information need is already clear from their question, output `NONE'. Otherwise, output the numbered list of interpretations as unambiguous questions and nothing else.

\colorbox{yellow!20}{User:}

\textbf{Question}\\
\{question\}
}
\end{prompt}

\begin{prompt}[title={Prompt \thetcbcounter: Interpretation Labeling}, label=prompt:interpretation-labeling]
{\ttfamily
\colorbox{cyan!20}{System:} You are an expert discourse analyst. A Reddit answer segment has already been labeled with a discourse action. Your task is to determine which interpretation of the original question the segment best addresses.

\textbf{Rules}\\
1. You are given a question, its possible interpretations, and a segment from an answer that has been labeled with a discourse action.\\
2. Determine which question interpretation the segment most directly addresses, adopts, or targets. This may be explicit or implicit.\\
3. If the segment clearly and directly addresses one of the interpretations, return that interpretation's ID.\\
4. If the segment does not clearly target any specific interpretation, return ``NONE''.\\

\textbf{Output Format}\\
Respond with exactly ONE JSON object:\\
{[}\{"interpretation\_id": "id\_1"\}{]}\\

When no specific interpretation is targeted:\\
{[}\{"interpretation\_id": "NONE"\}{]}\\

\colorbox{yellow!20}{User:}

\textbf{Question}\\
\{question\}

\textbf{Question Interpretations}\\
\{interpretations\}

\textbf{Full Answer}\\
\{answer\}

\textbf{Segment} (labeled as ``\{action\_label\}'')\\
\{segment\}

Respond with EXACTLY ONE JSON dictionary (NOT an array).
}
\end{prompt}

\section{Community Mimicking}
\begin{prompt}[title={Prompt \thetcbcounter: Mimicking Community Guidelines}, label=prompt:mimic]
{\ttfamily
\colorbox{cyan!20}{System:} r/\{subreddit\} is a subreddit for \{subreddit\_explanation\}. The community guidelines for r/\{subreddit\} are as follows: \{community\_guidelines\}.

\colorbox{yellow!20}{User:} Answer the question as if you were a redditor in that subreddit: \{question\}
}
\end{prompt}

\subsection{r/AskHistorians Community Guidelines}
{\footnotesize
Answers in r/AskHistorians are held to a higher standard than is generally found on Reddit. Answers should be in-depth, comprehensive, accurate, and based off of good quality sources. Answers should be in line with the existing historiography on the topic, and written in a manner respecting the Historical Method.

\textbf{Write an in-depth answer.} An in-depth answer provides the necessary context and complexity that the given topic calls for, going beyond a simple cursory overview. ``Good answers aren't good just because they are right --- they are good because they explain.'' Your answer should give context to the events being discussed, not simply list related facts. Ask yourself: Do I have the expertise needed to answer this question? Have I done research on this topic? Can I cite academic quality primary and secondary sources? Can I answer follow-up questions?

\textbf{Sources.} Answers should reflect current academic understanding or debates on the subject. Both primary and secondary sources are accepted. Secondary literature should be from academic or respected general publishers. Tertiary sources such as Wikipedia may be cited for basic undisputed facts, but sole reliance on tertiary sources is not allowed. Blog posts and random web articles are not acceptable. You are not a source: personal experience and anecdotes are not acceptable.

\textbf{No personal anecdotes.} Personal anecdotes are unreliable, unverifiable, and of very little real interest.

\textbf{No speculation.} Suppositions and personal opinions are not a suitable basis for an answer. Warning phrases include ``I guess,'' ``I believe,'' ``I think,'' ``to my understanding,'' and ``it makes sense to me that.''

\textbf{No partial answers or placeholders.} An answer should be full and complete in and of itself. Do not post partial answers with the intention of prompting further discussion or as a placeholder to expand on later.

\textbf{No political agendas or moralising.} Answers should represent a sincere effort to make an argument from the historical record, constructed in keeping with the principles of the historical method. Evidence should not be chosen selectively to support a predetermined argument.

\textbf{Do not just post links or quotations.} A good answer is a balanced mix of context, explanation, and sources. Always provide context for any source you cite. Answers should not consist only or primarily of copy-pasted text.

\textbf{No plagiarism.} Directly copying and pasting another person's words and passing them off as your own will result in an instant ban.
}

\subsection{r/ScienceBasedParenting Community Guidelines}
{\footnotesize
\textbf{Be respectful.} Discussions and debates are welcome but must remain civilized. Inflammatory content is prohibited.

\textbf{Read linked material before commenting.} Make sure you know what you are commenting on to avoid misunderstandings.

\textbf{Respect post flair.} All top-level comments must adhere to the flair type guidelines set by the OP. If you reply to a top-level comment with additional or conflicting information, a link to flair-appropriate material is also required.

\textbf{Sources must match flair type.} For ``Question --- Link To Research Required'' flair, top-level answers must link directly to peer-reviewed research published in scientific journals. For ``Question --- Link to Expert Consensus Required'' flair, links to sources containing expert consensus are permitted (e.g.\ CDC, WHO, American Academy of Pediatrics). Parenting books, podcasts, and blogs are not peer-reviewed and should not be referenced as scientific sources.

\textbf{No individualized medical advice.} General questions are allowed; specific questions about one's own child's symptoms or treatment are not. Nothing posted constitutes medical advice.

\textbf{Keep comments relevant.} All threads and comments must be directly relevant to the discussion. Off-topic threads and comments will be removed.

\textbf{No self-promotion or product promotion.} Do not use this subreddit to advertise or sell a product, service, podcast, or book.

\textbf{General discussion.} Questions that cannot be answered by direct research or expert consensus, or requests for anecdotes and parent-to-parent advice, belong in the weekly General Discussion Megathread.
}

\section{Computational Resources}
All experiments were run on two NVIDIA RTX
A6000s. Models greater than 11B parameters were
run using 4 bit quantization.
All models including closed-source models were run with a temperature of 0.01.

\begin{figure*}
\centering
\includegraphics[scale=0.5]{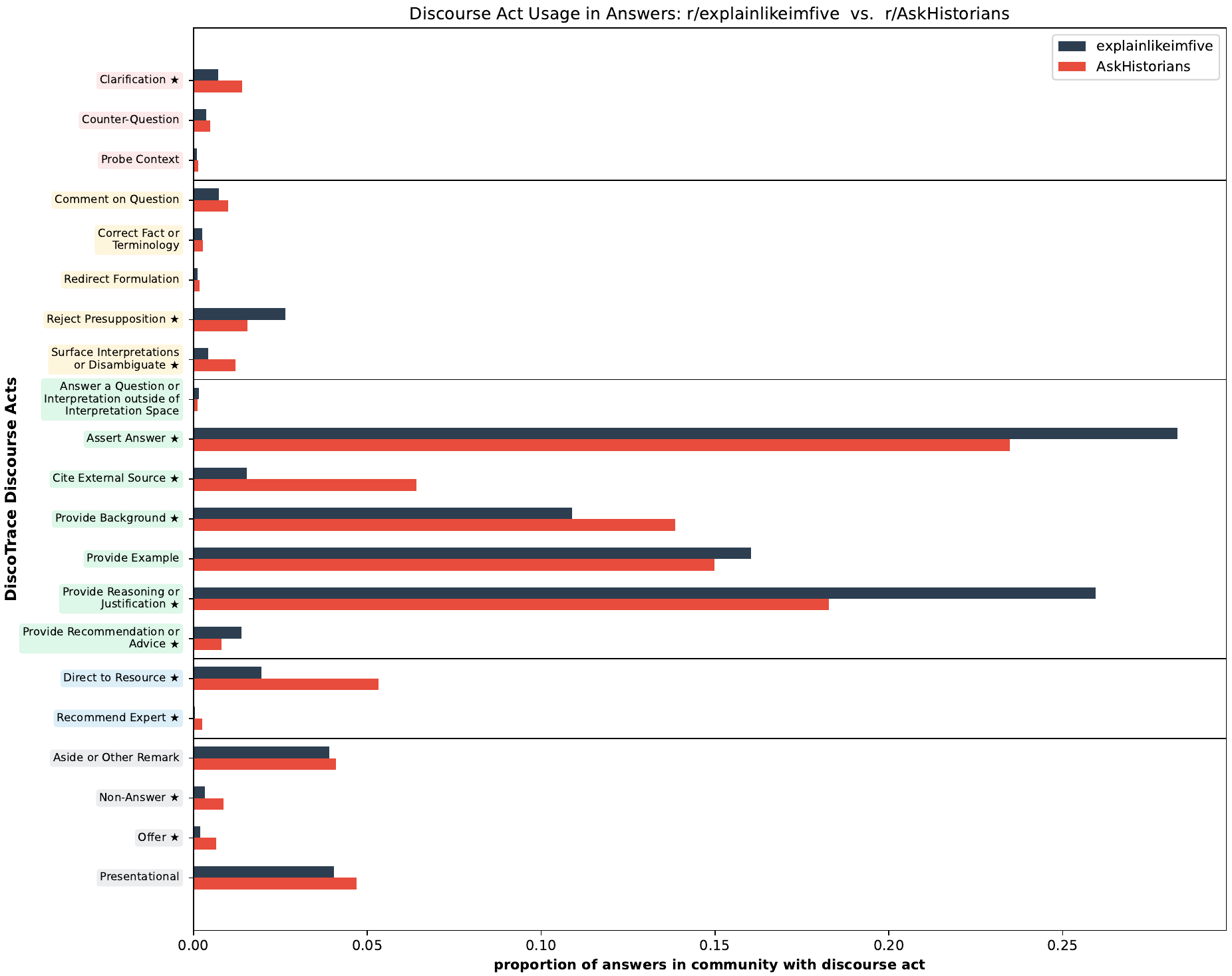}
\caption{We compare the proportion of answers that address each of the 21 acts in our ontology in \subreddit{explainlikeimfive} and \subreddit{AskHistorians}, revealing interesting differences in their act usage. Proportions differences that are statistically significant are starred, and measured using a chi-squared test with Bonferroni correction.}
\label{fig:act-comparison-eli5-v-askhistorians}
\end{figure*}

\begin{figure*}[t]
\centering
\includegraphics[scale=0.5]{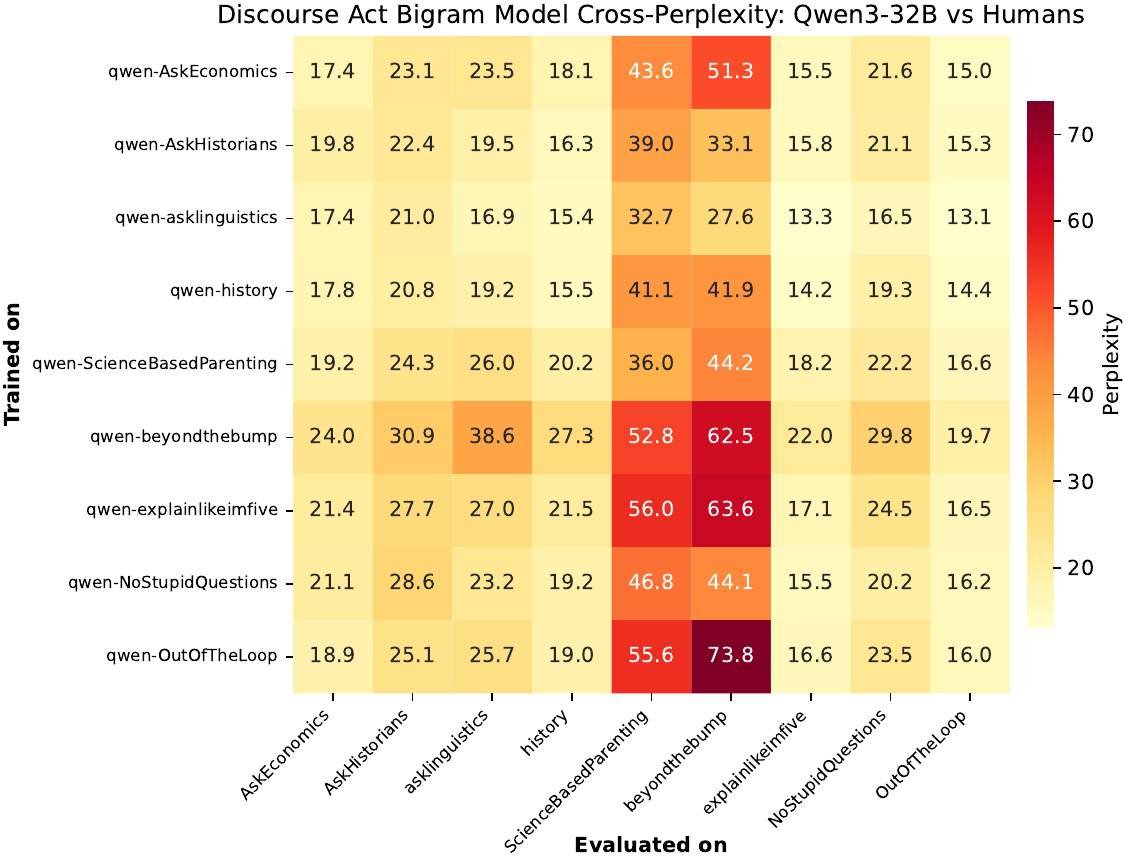}
\caption{We train bigram models on \textit{act} sequences from one community $C_\text{train}$ (rows) and evaluate on another $C_\text{eval}$ (columns).  The diagonal compares the LLM's answers against human answers in the same community. Here we train a bigram model on \model{qwen3-32b} answers to questions in $C_\text{train}$ and evaluate on human answers.}
\label{fig:qwen-v-human}
\end{figure*}

\begin{figure*}[t!]
\centering
\includegraphics[scale=0.5]{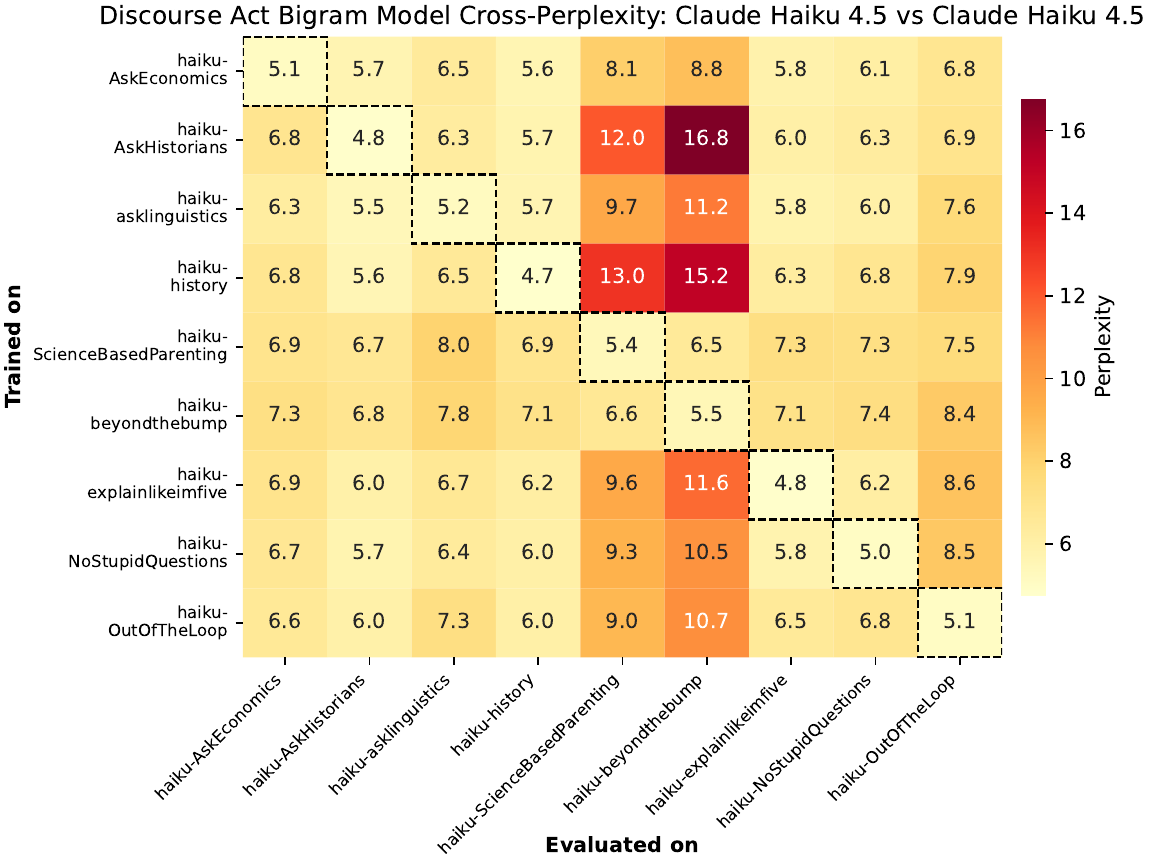}
\caption{We train bigram models on \textit{act} sequences from one community $C_\text{train}$ (rows) and evaluate on another $C_\text{eval}$ (columns).  The diagonal is the self perplexity. Here we train a bigram model on \model{claude-4.5-haiku} answers to questions in $C_\text{train}$ and evaluate on \model{claude-4.5-haiku} answers.}
\label{fig:haiku-v-haiku}
\end{figure*}

\begin{figure*}[t!]
\centering
\includegraphics[scale=0.6]{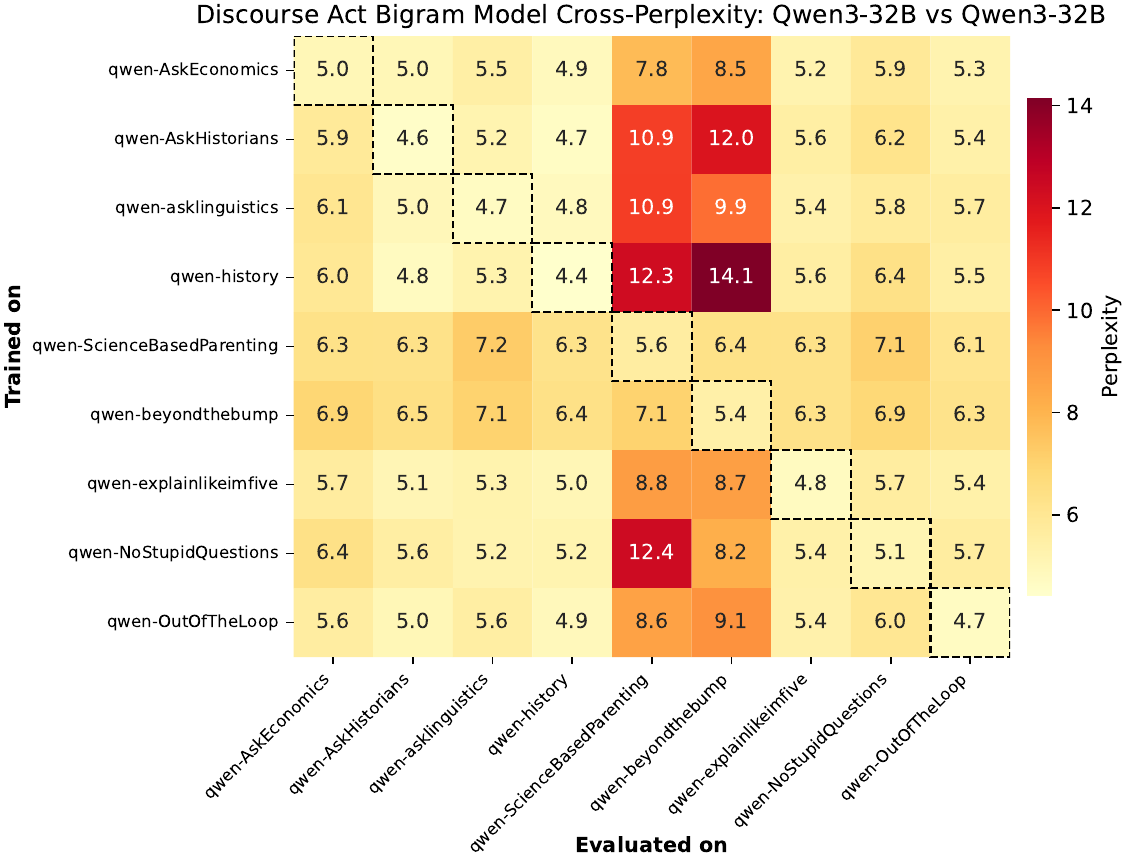}
\caption{We train bigram models on \textit{act} sequences from one community $C_\text{train}$ (rows) and evaluate on another $C_\text{eval}$ (columns).  The diagonal is the self perplexity. Here we train a bigram model on \model{qwen3-32b} answers to questions in $C_\text{train}$ and evaluate on \model{qwen3-32b} answers.}
\label{fig:qwen-v-qwen}
\end{figure*}

\begin{table}[t]
\centering
\resizebox{\columnwidth}{!}{%

\begin{tabular}{lccccc}
\toprule
Community & $|S_I|$ & Unmatched & \# Interps & Coverage \\
 & $\mu \pm \sigma$ & rate & $\mu \pm \sigma$ & $\mu \pm \sigma$ \\
\midrule
r/explainlikeimfive & 3.1$\pm$1.4 & 0.19 & 1.3$\pm$0.8 & 0.45$\pm$0.25 \\
r/AskHistorians & 3.3$\pm$1.7 & 0.25 & 1.3$\pm$0.9 & 0.41$\pm$0.27 \\
r/beyondthebump & 3.2$\pm$1.5 & 0.24 & 1.2$\pm$0.7 & 0.41$\pm$0.24 \\
r/ScienceBasedParenting & 3.1$\pm$1.5 & 0.32 & 1.2$\pm$0.8 & 0.39$\pm$0.25 \\
r/asklinguistics & 3.1$\pm$1.6 & 0.19 & 1.2$\pm$0.7 & 0.39$\pm$0.25 \\
r/OutOfTheLoop & 3.4$\pm$1.6 & 0.20 & 1.1$\pm$0.7 & 0.36$\pm$0.24 \\
r/history & 3.8$\pm$2.0 & 0.23 & 1.2$\pm$0.8 & 0.35$\pm$0.24 \\
r/NoStupidQuestions & 3.6$\pm$1.6 & 0.17 & 1.1$\pm$0.7 & 0.34$\pm$0.22 \\
r/AskEconomics & 3.9$\pm$1.9 & 0.25 & 1.3$\pm$0.9 & 0.34$\pm$0.24 \\
\bottomrule
\end{tabular}}
\caption{Interpretation space statistics across communities. Coverage is computed over questions with $|S_I| \geq 2$.}
\label{tab:interp_stats}
\end{table}

\begin{table}
\centering
\resizebox{\columnwidth}{!}{%

\begin{tabular}{clcccc} 
\toprule
\textbf{prompt mode} & \textbf{answerer} & \textbf{segments} & \textbf{matched} & \multicolumn{2}{c}{\textbf{dedication}} \\ 
 & & \textbf{per answer} & \textbf{per answer} & \textbf{mean} & \textbf{median} \\
\cmidrule{1-1}\cmidrule{2-6}
--- & human & 3.3 & 2.5 & 0.62 & 0.53 \\ 
\midrule
\multirow{2}{*}{default} & \texttt{haiku-4.5} & 5.7 & 4.6 & 0.46 & 0.40 \\
 & \texttt{qwen3-32b} & 12.1 & 10.4 & 0.40 & 0.31 \\ 
\midrule
default & \texttt{sonnet-4.5}* & 5.8 & 4.8 & 0.48 & 0.40 \\
mimic & \texttt{sonnet-4.5}* & 9.8 & 8.0 & 0.41 & 0.33 \\
\bottomrule
\end{tabular}}
\caption{Average number of eligible segments per answer, segments matched to an interpretation, and proportion of eligible segments dedicated to each addressed interpretation. Higher dedication values indicate more focused answers.}
\label{tab:interpretation-dedication-long}
\end{table}

\end{document}